\documentclass[letterpaper]{article} 
\usepackage[]{aaai2026}  
\usepackage{times}  
\usepackage{helvet}  
\usepackage{courier}  
\usepackage[hyphens]{url}  
\usepackage{graphicx} 
\usepackage{natbib}  
\usepackage{caption} 
\usepackage{algorithm}
\usepackage{algorithmic}

\usepackage{newfloat}
\usepackage{listings}
\DeclareCaptionStyle{ruled}{labelfont=normalfont,labelsep=colon,strut=off} 
\floatstyle{ruled}
\newfloat{listing}{tb}{lst}{}
\floatname{listing}{Listing}
\usepackage{amsmath,amssymb}
\usepackage{multirow}
\usepackage{colortbl}
\usepackage{color}
\usepackage{booktabs}
\usepackage{graphicx}
\usepackage{pifont}
\usepackage{bbding}
\usepackage{xcolor}
\usepackage[dvipsnames]{xcolor}
\usepackage{colortbl}
\usepackage[table]{xcolor}
\usepackage{enumitem}

\usepackage{subcaption}
\usepackage{caption}
\nocopyright

\def\eg{\textit{e.g.},~}               
\def\ie{\textit{i.e.},~}               
\def\vs{\textit{vs.}~}                 
\def\wrt{\textit{w.r.t.}~}              

\def\cuebench{\textsc{CueBench}}
\def\cue{\textsc{Cue}}
\def\NA{\textcolor{gray}{\textit{NA}}}
\def\YES{\textcolor{SeaGreen}{\ding{51}}}
\def\NO{\textcolor{Red}{\ding{55}}}
\def\HALFYES{\textsuperscript{\textcolor{Red}{\kern+0.8em\ding{55}}}\textcolor{SeaGreen}{\kern-0.8em\ding{51}}}

\def\Attribute{\textcolor{YellowOrange}{\textit{Attribute}}}

\newlength\paramargin
\newlength\figmargin

\newlength\secmargin
\newlength\figcapmargin
\newlength\tabcapmargin

\newcommand{\secref}[1]{Section~\ref{sec:#1}}
\newcommand{\figref}[1]{Figure~\ref{fig:#1}} 
\newcommand{\tabref}[1]{Table~\ref{tab:#1}}
\newcommand{\appenref}[1]{Appendix \S ~\ref{appen:#1}}

\newcommand{\algref}[1]{Algorithm~\ref{#1}}

\long\def\ignorethis#1{}

\newbox\jsavebox%

\title{\includegraphics[width=1.5em]{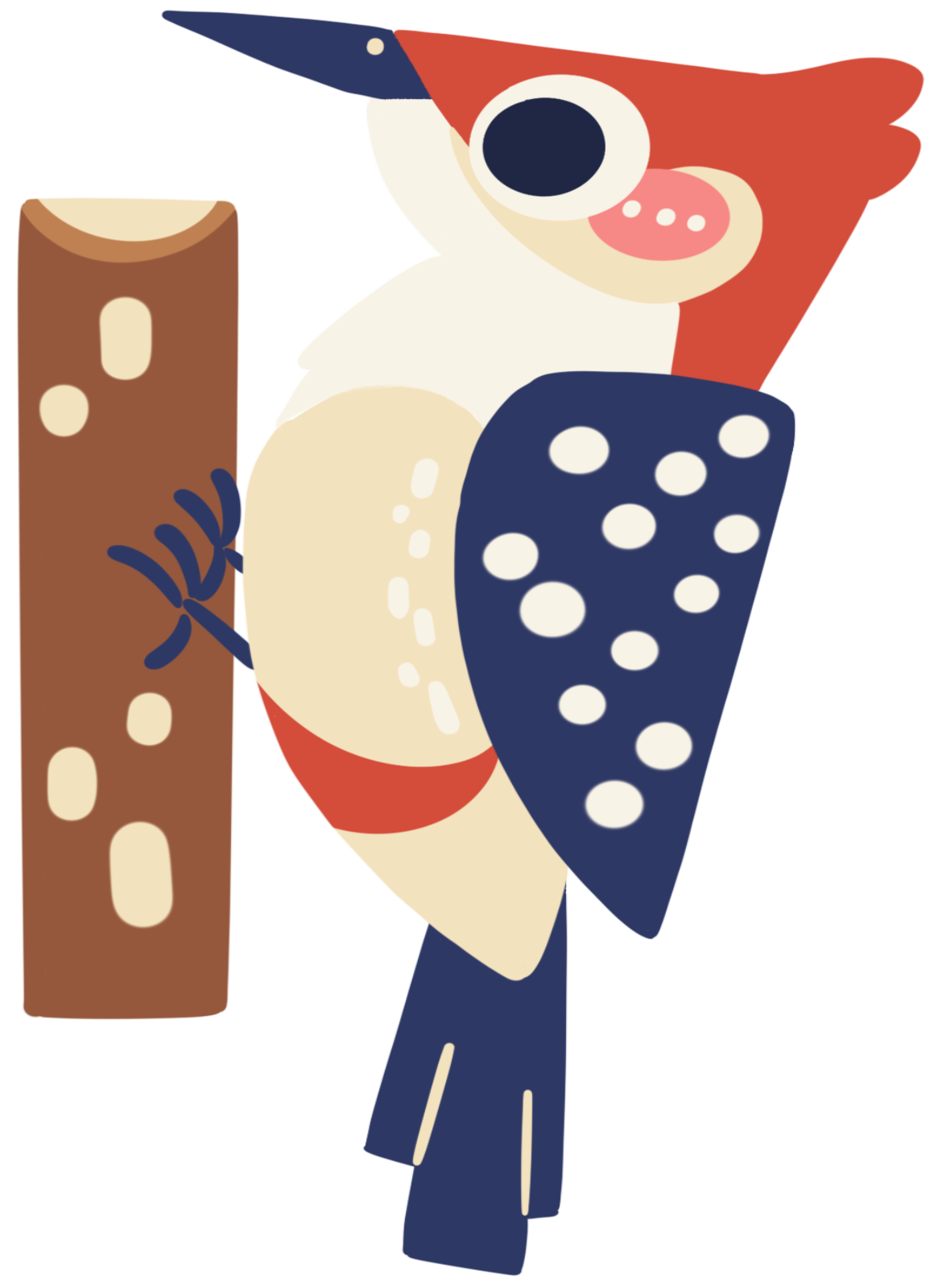}\cuebench: Advancing Unified Understanding of Context-Aware\\Video Anomalies in Real-World}
\author{
    Yating Yu\equalcontrib, 
    Congqi Cao\equalcontrib\thanks{Corresponding author.}, 
    Zhaoying Wang, 
    Weihua Meng, \\ 
    Jie Li, 
    Yuxin Li, 
    Zihao Wei, 
    Zhongpei Shen, 
    Jiajun Zhang
}
\affiliations{
    Northwestern Polytechnical University, Xi'an Shaanxi, 710129, China\\
    yatingyu@mail.nwpu.edu.cn, congqi.cao@nwpu.edu.cn
}

\begin{document}

\maketitle

%
\begin{abstract}
How far are deep models from real-world video anomaly understanding (VAU)?
Current works typically emphasize detecting unexpected occurrences deviating from normal patterns or comprehending anomalous events with interpretable descriptions.
However, they exhibit only a superficial comprehension of real-world anomalies, with limited breadth in complex principles and subtle context that distinguish the anomalies from normalities, \eg climbing cliffs with safety gear \vs without it. 
To this end, we introduce \textbf{\cuebench}, the first of its kind \underline{Bench}mark, devoted to \underline{C}ontext-aware video anomalies within a \underline{U}nified \underline{E}valuation framework. 
We comprehensively establish an event-centric hierarchical taxonomy that anchors two core event types: $14$ conditional and $18$ absolute anomaly events, defined by their refined semantics from diverse contexts across $174$ scenes and $198$ attributes.
Based on this, we propose to unify and benchmark context-aware VAU with various challenging tasks across recognition, temporal grounding, detection, and anticipation. 
It also serves as a rigorous and fair probing evaluation suite for generalized and specialized vision-language models (VLMs) across both generative and discriminative paradigms.
To address the challenges underlying \cuebench, we further develop \textbf{\cue-R1} based on R1-style reinforcement fine-tuning with verifiable, task-aligned, and hierarchy-refined rewards in a unified generative manner. 
Extensive results on \cuebench~reveal that, existing VLMs are still far from satisfactory real-world anomaly understanding, while our \cue-R1 surpasses these state-of-the-art approaches by over $24\%$ on average.
%
\end{abstract}
\begin{links}
    \link{Code}{https://github.com/Mia-YatingYu/Cue-R1}
    \link{Datasets}{https://huggingface.co/datasets/CueBench/CueBench}
\end{links}

\begin{figure*}[!t]
    \centering
         \setlength{\abovecaptionskip}{4pt}
    \setlength{\belowcaptionskip}{0pt}
    \includegraphics[width=\textwidth]{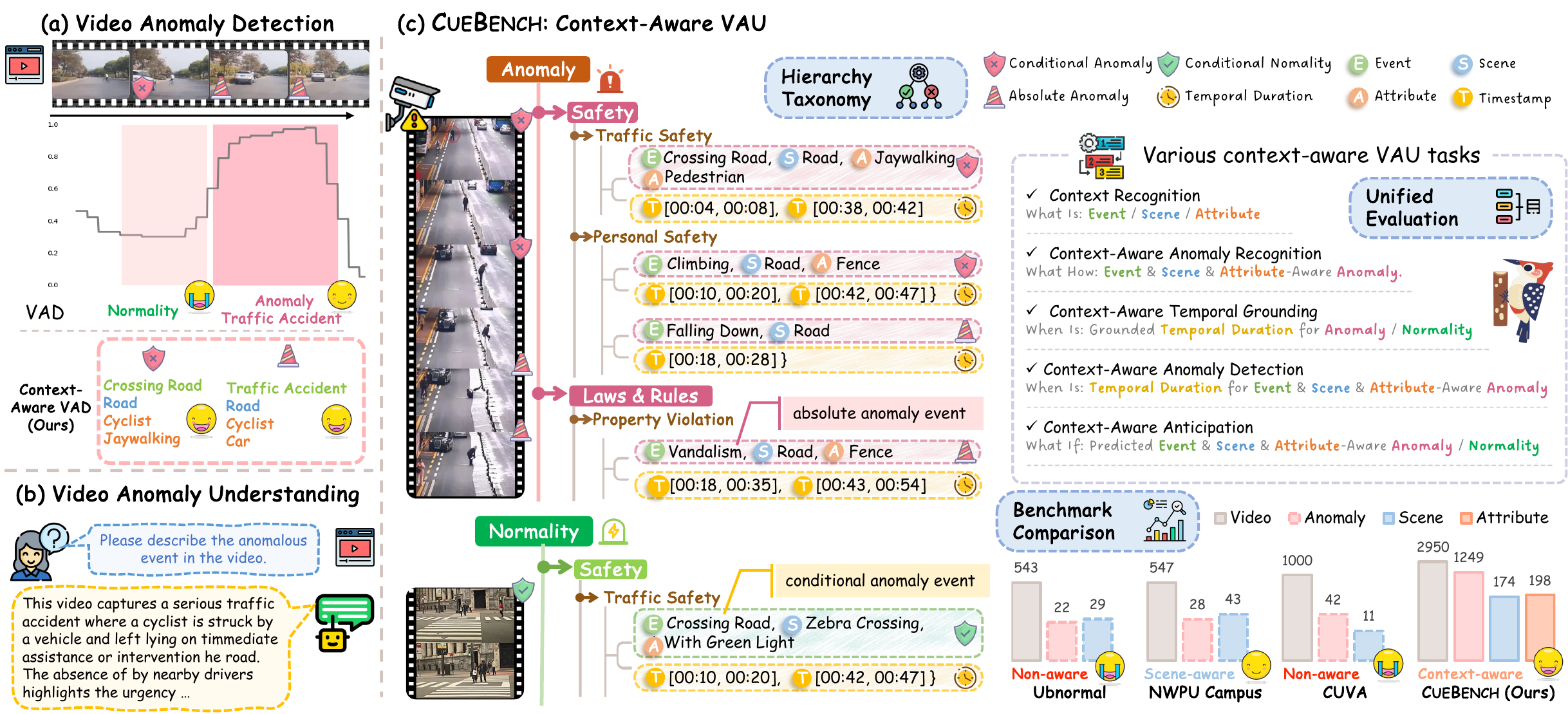}
    \caption{\textbf{Comparison of existing benchmarks.} (a) Traditional VAD aims to detect deviations from normal patterns and identify the time window of the occurring anomaly, yet exhibiting insufficient comprehension of subtle anomalies and lacking context-awareness (\eg \textit{cyclist jaywalking while crossing road}). (b) Current VAU benchmarks primarily emphasize the interpretation of absolutely anomalous events with explainable outputs. (c) Our large-scale \cuebench~features a diverse collection of \textbf{context-aware anomalies and normalities} from real-world scenarios, organized within a comprehensive \textbf{hierarchical taxonomy}, and supports \textbf{unified evaluation} across five challenging VAU tasks. }
    \label{fig:cuebench} 
\end{figure*}
\section{Introduction}
Video anomaly understanding (VAU) derived from general video understanding, emphasizes the automated comprehension of anomalous events in videos, which encompasses a diverse range of tasks including anomaly detection~\cite{ristea2024self,cai2021appearance,cao2024context,yan2023feature}, recognition~\cite{wu2024vadclip,yu2025building}, and localization~\cite{zhou2016spatial}.
At its core, video anomaly detection (VAD) tends to detect deviations from the learned normal patterns~\cite{zhu2022towards}.
Keeping pace with the development of VLMs~\cite{radford2021learning,bai2025qwen2,yu2025unhackable,zhang2025videollama,comanici2025gemini,hurst2024gpt}, a growing body of works has emerged to comprehend anomalies in open-vocabulary settings~\cite{wu2024open,li2025anomize,zanella2024harnessing} and further in a VQA manner with interpretable explanations~\cite{du2024uncovering,zhang2025holmes,ye2025vera,xu2025towards,du2024exploring,ma2025sherlock,huang2025vad}.
Given that real-world anomalies are complex, diverse, and evolving, there is a need for a more \textbf{realistic} and \textbf{universal} comprehension that aligns with human experiences and societal norms.
With current advancements, a natural question raises: \textit{How far are current VLMs from truly understanding of real-world video anomalies?}

While existing works are appealing, they oversimplify the nature of real-world anomalies. 
Some studies have explored the role of contextual semantics in VAU~\cite{wu2024open,ma2025sherlock}, but their focus has largely been on comprehending traditional \textit{absolute anomaly events} (\eg ``explosion'', ``car crash'') or simple \textit{deviations} (\eg ``biking'' instead of the expected ``walking''), where contextual cues are not decisive in determining normality \vs anomaly.
Recent efforts have drawn attention to scene dependencies underlying anomalies~\cite{cao2023new,cao2025scene,zhang2025autoregressive}, yet the reliance on scene-only contexts and the sparsity of scene-dependent anomalies reveal a substantial gap in real-world VAU.
%
%
%
In practice, the same event (\eg ``climbing'') could be interpreted as normal or abnormal depending on both scene and attribute context: ``climbing cliffs with safety gear'' is normal, whereas ``climbing cliffs without any protection'' is clearly abnormal, due to the inherent risks in cliff scenes and the need for additional precaution.
Such \textit{conditional anomaly events}, implying ambiguous boundaries and subtle context dependencies from both scenes and attributes, remain largely underexplored in existing works. 

For a long period, VAU research has followed task-specific paradigms, designing specialized architectures and loss functions to cater to unique requirements of separate tasks and benchmarks.
Despite the breadth, VAU has predominantly focused on specific capabilities like VAD, multi-modal retrieval and VQA, leading to fragmented and incompatible solutions.
Such fragmentation underscores the need for a unified framework benchmarking diverse demands, fostering holistic and integrated real-world VAU.

To satisfy these desiderata, we develop \textbf{\cuebench}, the first benchmark dedicated to unified, context-aware video anomaly understanding in real-world.
Compared with existing benchmarks in~\figref{cuebench}, \cuebench~highlights the following distinct characteristics:
\begin{itemize}
    \item \textbf{Context Awareness.}
    Given the complex context dependencies of real-world anomalies, \cuebench~is the first to introduce and integrate the concepts of $18$ \textit{absolute anomaly events} (\eg ``falling down'', ``vandalism'') and $14$ \textit{conditional anomaly events} (\eg ``crossing road'', ``climbing'') \wrt subtle contextual cues drawn from $174$ scenes and $198$ attributes. 
    Note that both anomalies and normalities in \cuebench~are represented as context triplets comprising events, scenes, and attributes.
    Hence, along with the diversity of anomalies ($1249$), the normalities ($194$) are context-dependent and diversified as well, going beyond the rare occurrences and monotonous absolute anomalies in existing benchmarks.
    
    
    \item \textbf{Comprehensive Hierarchical Taxonomy.}
    As anomalies vary widely in types, contexts, and impacts, we comprehensively build a 5-level event-centric hierarchical taxonomy, extending from the fundamental anomaly \vs normality to fine-grained triplets. The key insight is that anomaly errors often carry far more severe consequences than simple context misinterpretations. The refined differentiation of violation and inherent severity (\eg on \textit{safety}, \textit{laws\&rules}, \textit{life\&health}) in the hierarchy enables trustworthy evaluation and prioritization in real-world.

    \item \textbf{Unified Evaluation Framework.}
    In contrast to existing VAU benchmarks focusing on task-specific paradigm, \cuebench~distinguishes itself by adopting a unified generative evaluation framework, where VLMs are prompted with videos and task-specific queries. 
    Through a suite of five test tasks and crafted evaluation metrics, \cuebench~enables comprehensive gauge of models' capabilities across recognition, detection, grounding, and anticipation.
    Within the unified task space, we hope that further development of universal architectures and training objectives will continue to advance this field.
\end{itemize}

Leveraging this dataset, we present \textbf{\cue-R1}, a unified generative approach that incorporates supervised and reinforcement fine-tuning with verifiable, task-aligned, and hierarchy-refined rewards tailored to context-aware VAU.
Extensive results on \cuebench~reveal that existing generalized and specialized VLMs, both generative and discriminative, remain unsatisfactory, while \cue-R1 provides new insights for developing a universal solution.

\begin{table*}[!htbp]
\centering
\setlength{\abovecaptionskip}{5pt}
\setlength{\belowcaptionskip}{-7pt}
\setlength{\tabcolsep}{1mm}
\resizebox{1\textwidth}{!}
{
\begin{tabular}{lcccccccccccccc}
\toprule
\multirow{2}{*}{\textbf{Benchmark}} & \multirow{2}{*}{\textbf{Domain}} & \multirow{2}{*}{\textbf{Length}} & \multirow{2}{*}{\textbf{\#Video}} & \multirow{2}{*}{\begin{tabular}[c]{@{}c@{}}\textbf{\#Absolute}\\ \textbf{Anomaly}\end{tabular}} & \multirow{2}{*}{\begin{tabular}[c]{@{}c@{}}\textbf{\#Conditional}\\ \textbf{Anomaly}\end{tabular}} & \multirow{2}{*}{\textbf{\#Normality}} & \multirow{2}{*}{\begin{tabular}[c]{@{}c@{}}\textbf{Anomaly}\\ \textbf{Dependency}\end{tabular}} & \multirow{2}{*}{\textbf{\#Scene}} & \multirow{2}{*}{\textbf{\#Attribute}} &  \multicolumn{5}{c}{\textbf{Task Setup}} \\ \cmidrule{11-15}
 &   &  &   &  &  &  &  &   &  & \textbf{R} & \textbf{G} & \textbf{D} & \textbf{A} & \textbf{Q} \\
\cmidrule{1-15}
\rowcolor{gray!15} 
\multicolumn{15}{c}{\textbf{\textit{Traditional Video Anomaly Detection Datasets}}} \\
Subway Entrance~\cite{adam2008robust}  & Pedestrian & 1.5h & 1 & 5 & \NA & 1 & Deviation & 1 & \NA &  \NO & \NO & \YES & \NO & \NO \\
Subway Exit~\cite{adam2008robust}  & Pedestrian & 1.5h & 1 & 3 & \NA & 1 & Deviation & 1 & \NA &  \NO & \NO & \YES & \NO & \NO \\
UCSD Ped1~\cite{wang2010anomaly} & Pedestrian & 0.1h & 5 & 5 & \NA & 1 & Deviation & 1 & \NA &  \NO & \NO & \YES & \NO & \NO \\
UCSD Ped2~\cite{wang2010anomaly}  & Pedestrian & 0.1h & 5 & 5 & \NA & 1 & Deviation & 1 & \NA &  \NO & \NO & \YES & \NO & \NO \\
CUHK Avenue~\cite{lu2013abnormal}  & Pedestrian & 0.5h & 5 & 5 & \NA & 1 & Deviation & 1 & \NA &  \NO & \NO & \YES & \NO & \NO \\
ShanghaiTech~\cite{luo2017revisit} & Pedestrian & - & 13 & 11 & \NA & 1 & Deviation & 13 & \NA &  \NO & \NO & \YES & \NO & \NO \\
UCF-Crime~\cite{sultani2018real}  & Crime & 128h & 1900 & 13 & \NA & 1 & Event & \NA & \NA &  \NO & \NO & \YES & \NO & \NO \\
Street Scene~\cite{ramachandra2020street}  & Traffic & 3.7h & 81 & 17 & \NA & 1 & Deviation & 1 & \NA &  \NO & \NO & \YES & \NO & \NO \\
XD-Violence~\cite{wu2020not}  & Violence & 217h & 4754 & 6 & \NA & 1 & Event & \NA & \NA &  \NO & \NO & \YES & \NO & \NO \\
Ubnormal~\cite{acsintoae2022ubnormal}  & Pedestrian & 2.2h & 543 & 22 & \NA & 1 & Deviation & 29 & \NA &  \NO & \NO & \YES & \NO & \NO \\
NWPU Campus~\cite{cao2023new}  & Pedestrian & 16h & 547 & 28 & 4 & 1 & Event, Scene & 43 & \NA &  \NO & \NO & \YES & \YES & \NO \\
MSAD~\cite{zhu2024advancing}  & Multiple & - & 720 & 55 & \NA & 1 & Event & 14 & \NA &  \NO & \NO & \YES & \NO & \NO \\
\midrule
\rowcolor{gray!15} 
\multicolumn{15}{c}{\textbf{\textit{Video Anomaly Understanding Datasets}}} \\
CUVA~\cite{du2024uncovering}  & Multiple & 32.5h & 1000 & 42 & \NA & \NA & Event & 11 & \NA &  \YES & \NO & \YES & \NO & \YES \\
HAWK~\cite{tang2024hawk}  & Mixture & 142.5h & 8000 & - & \NA & \NA & Event & \NA & \NA &  \NO & \NO & \NO & \NO & \YES \\
HIVAU-70k~\cite{zhang2025holmes}  & Mixture & - & 5443 & 19 & \NA & 1 & Event & \NA & \NA & \NO & \NO & \YES & \NO & \YES \\
\rowcolor{red!10} 
{\begin{tabular}[l]{@{}l@{}}\textbf{\cuebench}\\ \textbf{(Ours)}\end{tabular}}  & Multiple & 54.5h & 2950 & 18 $\rightarrow$ 840 & 14 $\rightarrow$ 409 & 14 $\xrightarrow{}$ 194 & {\begin{tabular}[c]{@{}c@{}}Event, Scene,\\Attribute\end{tabular}} & 174 & 198 &  \YES & \YES & \YES & \YES & \YES \\ \bottomrule
\end{tabular}
}
\caption{We review existing VAD and VAU benchmarks and highlight key characteristics of \cuebench. 
``Mixture'' denotes the combination of existing public datasets.
Different from others, \cuebench~is the first large-scale benchmark for context-aware VAU. 
Due to anomaly dependencies of contexts from absolute and conditional anomaly events with different scenes and attributes, \#Anomaly and \#Normality are highly diversified, progressing from event categories (L-4) to ($\rightarrow$) context triplets (L-5) in hierarchy taxonomy. 
It is designed to evaluate various tasks including anomaly recognition (R), temporal grounding (G), anomaly detection (D) and context anticipation (A), all of which can be approached in a unified VQA manner (Q).}
\label{tab:bench}
\end{table*}
\begin{figure*}[!t]
    \centering
    \setlength{\abovecaptionskip}{4pt}
    \setlength{\belowcaptionskip}{0pt}
    \includegraphics[width=\textwidth]{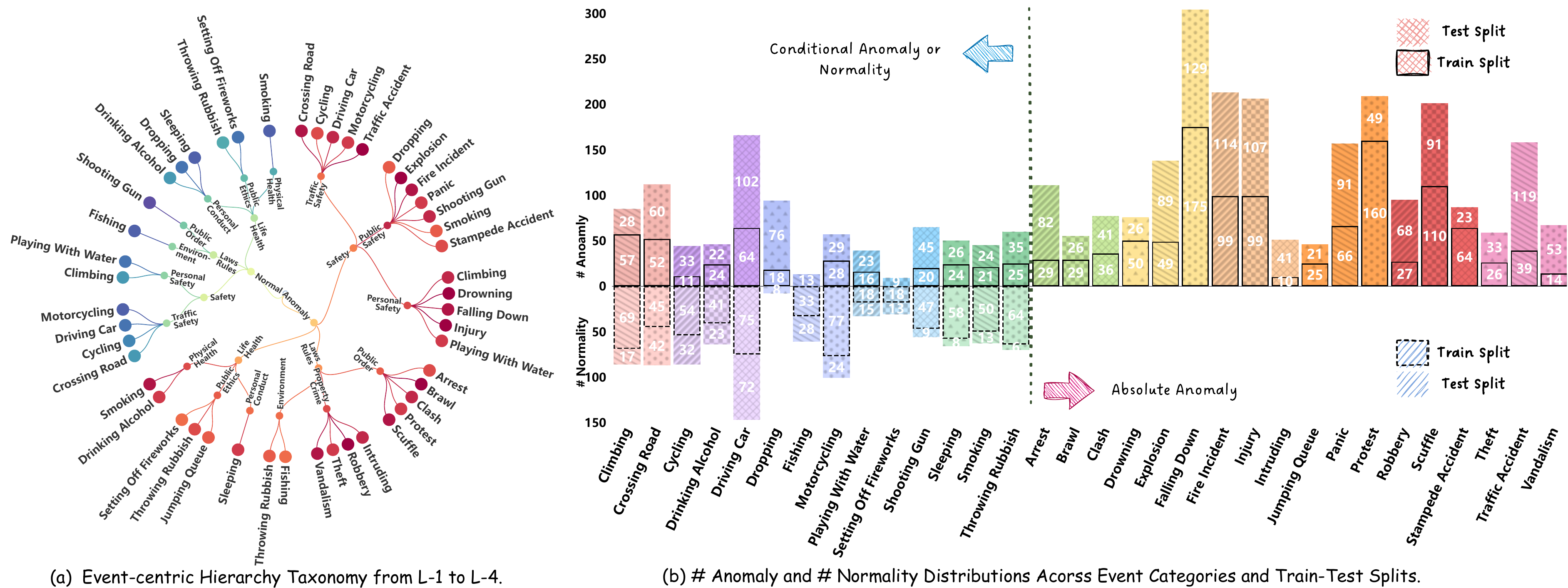}
    \caption{\textbf{Data statistics of \cuebench.} (a) We comprehensively build an event-centric hierarchical taxonomy that covers 2 states, 3 domains, and 9 effects in a top-down manner. And (b) our \cuebench~exhibits a diverse spectrum of conditional anomalies, normalities, and absolute anomalies across both the training and test splits.}
    \label{fig:statistics} 
\end{figure*}

%
%
\section{Benchmark: \text{C\small{UE}}\text{B\small{ENCH}}}

\subsection{Data Statistics}
\label{sec:data}
We compare \cuebench~with existing VAU benchmarks in~\tabref{bench}. 
Generally, our \cuebench~comprises 2,950 newly collected videos sourced from multiple domains on YouTube totaling 54.5 hours of footage. 
Each video ranges from 10s to 5min in length, with rich annotations of contexts and anomaly labels. The labeled context-aware segments span approximately 62\% of the total duration.

\noindent \textbf{Context Indispensability.}
Unlike existing VAU benchmarks, \cuebench~is explicitly designed to be context-indispensable, encompassing 174 scenes and 198 attributes besides 32 event categories \wrt 18 \textit{absolute anomaly events} and 14 \textit{conditional anomaly events}. 
Notably, it features 1,443 distinct context triplets, each representing a combination of an event with various scenes and attributes, linked to either an anomaly or normality.
This yields \textbf{840 absolute anomalies} (\eg $\langle$\texttt{vandalism, road, fence}$\rangle$), \ie triplets involving \textit{absolute anomaly events} that remain anomalous across various scenes and attributes, \textbf{409 conditional anomalies} (\eg $\langle$\texttt{crossing road, road, pedestrian jaywalking}$\rangle$), and \textbf{194 conditional normalities} (\eg $\langle$\texttt{crossing road, zebra crossing, green light}$\rangle$), \ie triplets containing \textit{conditional anomaly events} whose abnormal/normal states hinge on context cues.
Such rich contextual grounding ensures that understanding anomalies in \cuebench~requires nuanced reasoning beyond superficial event recognition.

\noindent \textbf{Event-centric Hierarchy Taxonomy.}
\cuebench~incorporates a comprehensive five-level event-centric hierarchy taxonomy, where each leaf node in Level 5 (L-5) represents a distinct context triplet \wrt a normality or anomaly. 
Due to the space limitations, we present the core top-4 levels of hierarchy in~\figref{statistics}(a), which captures a diverse spectrum of cognitive impacts in real-world.
At the top, L-1 distinguishes two fundamental states: Anomaly \vs Normality. 
This branches into three L-2 domains and further into nine L-3 effects underscoring both the shared and distinct characteristics across various real-world anomalies.
Note that L-4 comprises 34 event nodes, as certain conditional anomaly events \ie ``throwing rubbish'' and ``smoking'', exhibit two distinct anomaly effects depending on context.

\begin{figure*}[!t]
    \centering
    \setlength{\abovecaptionskip}{3pt}
    \setlength{\belowcaptionskip}{0pt}
    \includegraphics[width=\textwidth]{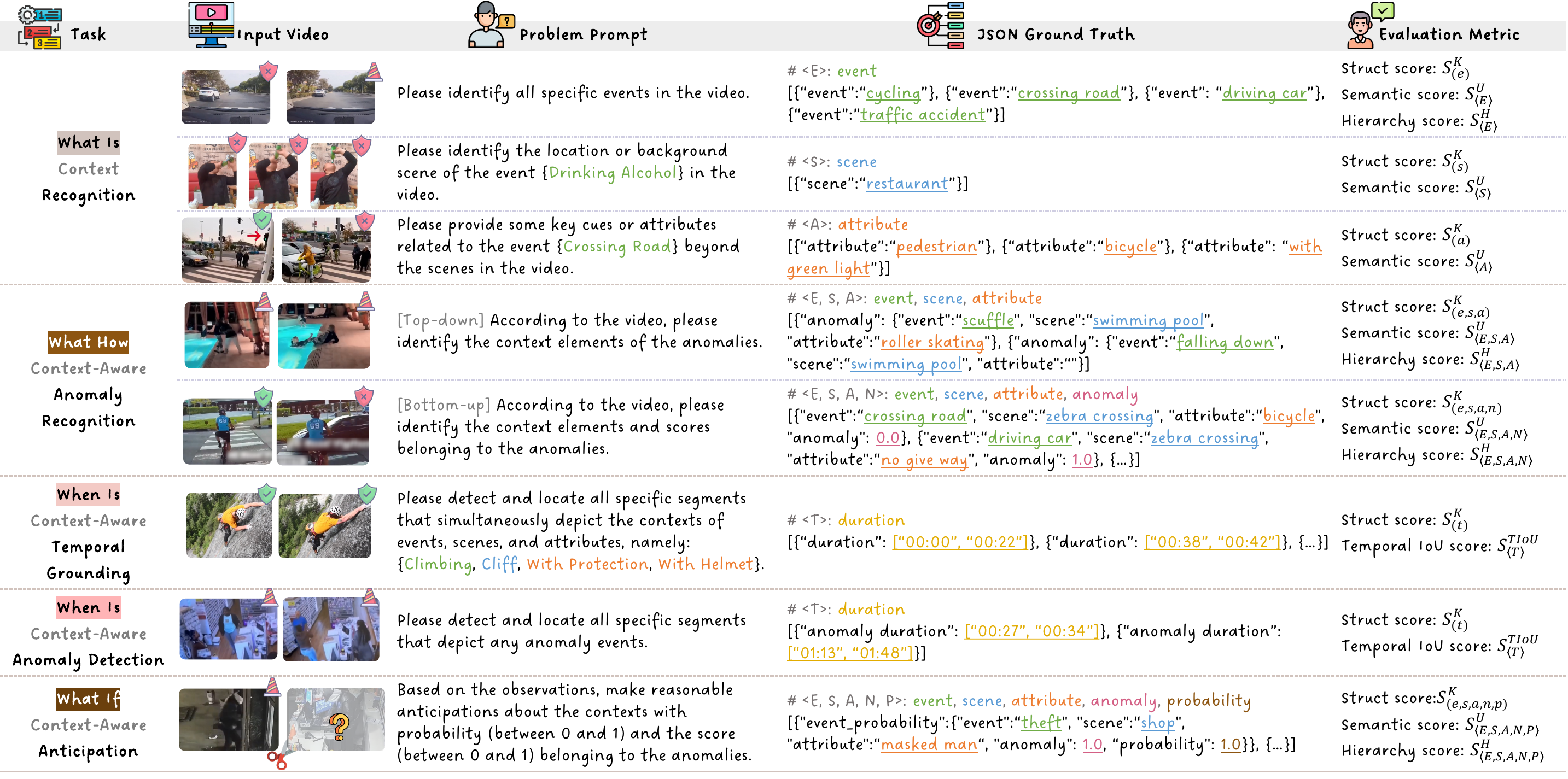}
    \caption{\textbf{Evaluation framework with task examples of \cuebench.} Our benchmark advances the evaluation of five challenging context-aware VAU tasks in a unified generative manner, by prompting the generative VLMs with videos and task-related problems. The VLMs are required to respond accordingly in a JSON-style format rather than free-texts. This enables accurate evaluation of various tasks for generative VLMs by checking the answers with ground-truths. }
    \label{fig:evaluation} 
\end{figure*}
\noindent \textbf{Training \& Testing Settings.}
To ensure an open-world setting, we divide the dataset into two sets: the test set comprising 1,222 videos covering all 1,443 distinct context triplets, while the training set with the remaining 1,728 videos containing only 440 context triplets. 
\figref{statistics}(b) presents the distributions of anomaly and normality across event categories and between the training and test splits of \cuebench.
Notably, the test set features higher density of context triplets than the training set (1.68 \vs 1.21 triplets per video), enabling a more challenging and realistic evaluation.

\subsection{Task Definition}
\label{sec:task}
To comprehensively evaluate models’ ability of VAU, we define a unified suite of five tasks built around the concept of context-aware reasoning. 
Each task targets a distinct yet complementary aspect of semantic, temporal and causal anomaly understanding, encouraging holistic perception and interpretation of video events under real-world complexity.

\noindent{\textbf{What Is.}} \textbf{(1)} \textit{Context Recognition}:
Identify the specific contextual elements (\ie events, scenes, or attributes) present in the video. This task serves as a fundamental perceptual evaluation of models' context-aware capabilities.

\noindent{\textbf{What How.}} \textbf{(2)} \textit{Context-Aware \textbf{Anomaly} Recognition.}
Identify the specific context triplets occurring in the video, and determine the existence of accurate absolute and conditional anomalies accordingly.
We introduce two paradigms for this task: \textbf{(a)} Automatically distinguish all anomalies from normalities with their corresponding contexts in a \textit{top-down} manner. \textbf{(b)} Extract the context triplets (whether anomalous or not), then assign anomaly scores to each group based on their semantics in a \textit{bottom-up} manner.

\noindent{\textbf{When Is.}} \textbf{(3)} \textit{Context-Aware Temporal Grounding.}
Ground target moments \ie one or more continuous intervals from untrimmed videos according to the queries based on context triplets that suggest an anomaly or a normality.
\textbf{(4)} \textit{Context-Aware \textbf{Anomaly} Detection.}
Automatically detect and localize all temporal clips that show any anomalies by ascertaining the contexts underlying the occurrences. 

\noindent{\textbf{What If.}}
\textbf{(5)} \textit{Context-Aware Anticipation.}
Infer the subsequent normalities or anomalies by reasoning the context triplets, based on the observed video clips.

\subsection{Evaluation Framework}
\label{sec:evaluation}
\figref{evaluation} presents the evaluation of five challenging context-aware VAU tasks in a unified generative manner.

\noindent {\textbf{Problem Formulation.}} Given a video input $\mathcal{V}$ along with the problem $\mathcal{T}_p$ and format prompt $\mathcal{T}_f$ \wrt task $\mathcal{T}$, we prompt generative VLMs to output the answer lists ($\mathcal{O}=[o_1,\dots,o_r]$) in a JSON format. 
According to $\mathcal{T}_p$ and $\mathcal{T}_f$, the model $\pi$ can generate different task-specific outputs.
The process can be formulated as:
\begin{equation}
    \{\mathcal{O},\mathcal{R}\} \text{ or } \{\mathcal{O}\}=\pi(\mathcal{V},\mathcal{T}_p,\mathcal{T}_f^K,\mathcal{T}_f^V),
\end{equation}
where $\mathcal{R}$ represents the response of the reasoning process, $\mathcal{T}_f^K$ and  $\mathcal{T}_f^V$ specify the required task-specific key names and value types respectively, \eg for bottom-up context-aware anomaly recognition, $\mathcal{T}_f^K=($event, scene, attribute, anomaly$)$, $\mathcal{T}_f^V=\langle E,S,A,N\rangle$.
Each element $o_i=\{o_i^K:o_i^V\}$ in $\mathcal{O}$ denotes a key-value pair, and the key bag and value content are formulated as $\mathcal{O}^K = \{o_i^K\}_{i=1}^{r}$ and $\mathcal{O}^V = \{o_i^V\}_{i=1}^{r}$, respectively.
This enables us to accurately probe various tasks by checking the VLM's output $\mathcal{O}$ with ground-truths $\mathcal{G}=[\{g_j^K:g_j^V\}_{j=1}^{t}]$ via a tailored evaluation metric suite to capture both structure alignment and task-related content quality, avoiding the bias scoring of LLMs.

\noindent {\textbf{Evaluation Metrics.}} To assess structure alignment, we design a structure-based F1 score which calculates binary matching between the output and ground-truth key bags $\mathcal{O}^K$ and $\mathcal{G}^K$ in the key space $K$:
\begin{equation}
    S^{K} = 
    \frac{2\left | \mathcal{O}^K\cap\mathcal{G}^K \right | }
    {2\left | \mathcal{O}^K\cap\mathcal{G}^K \right | + 
    \left | \mathcal{O}^K\setminus \mathcal{G}^K \right |+
    \left | \mathcal{G}^K \setminus \mathcal{O}^K \right |} ,
\end{equation}
For content quality evaluation of ``\verb|What|'' tasks, we first compute semantic embeddings~\cite{devlin2019bert} from value content of both output ($\mathcal{O}^V$) and ground-truth ($\mathcal{G}^V$), denoted as $\mathcal{O}^U$ and $\mathcal{G}^U$ in the Euclidean space $U$. 
We then construct a semantic matching matrix $\mathcal{M}\in \mathbb{R}^{r\times t}$, where each element $m_{i,j}$ is a binary variable indicating whether $o_i^U \in \mathcal{O}^U$ and $g_j^U\in\mathcal{G}^U$ are matched using Hungarian algorithm~\cite{kuhn1955hungarian} based on cosine similarity.
Thus, the semantic score $S^U$ is defined as:
\begin{equation}
   S^U=\frac{1}{r\cdot t}\sum_i\sum_j m_{i,j} \cdot \cos(o_i^U,g_j^U).
\end{equation}
Given that semantic similarity can be overly lenient to hallucinated answers and often fails to reflect task alignment accurately in anomaly understanding, we propose a novel hierarchy score.
This metric leverages the event-centric hierarchy taxonomy $H$ to better assess human-aligned performance for event-related tasks.
Unlike the semantic score, we retrieve the most likely leaf nodes of $o_i^U$ within $H$ as its proxy (anomaly or normality) $\hat{o}^H_i$, based on their semantic similarities, and $g_j^V$ can be reflected to $g_j^H$ directly. 
After that, the hierarchy distance $d^H_{i,j}$ of each paired proxy and ground truth $(\hat{o}_i^H,g_j^H)$ is computed and then normalized as the final hierarchy score:
\begin{equation}
 S^H = \frac{1}{r \cdot t} \sum_i \sum_j m_{i,j} \left(1 - \frac{d^H_{i,j}}{d^H_{\max}}\right) \cdot \mathbb{I}\left(d^H_{i,j} \leq \tau \cdot d^H_{\max} \right),
\end{equation} 
where $d_{\max}^{H}$ is the maximum depth of $H$ and $\tau$ is the threshold for valid hierarchy alignment.
For content quality evaluation of ``\verb|When|'' tasks, we adopt the temporal IoU metrics as the temporal score $S^\mathrm{TIoU}$. 


\section{Method: \cue-R1}
\label{sec:method}
To facilitate the comprehensive integration of context-aware capability \wrt various tasks into the training process, we develop \cue-R1 in a unified generative pipeline, based on reinforcement learning (RL) with GRPO algorithm~\cite{shao2024deepseekmath}. 
Following the rule-based reward paradigm of Open-R1~\cite{guo2025deepseek}, our RL setup requires reward signals that are both reliable and precise.
To ensure this, the training data is centered around tasks with clearly verifiable outputs, structured in a JSON format. 
This enables accurate reward computation using simple rules as mentioned in~\secref{evaluation}, thereby promoting stable and effective RFT~\cite{liu2025visual,shen2025vlm}.
Our rule-based accuracy reward seamlessly aligns the policy model $\pi_\theta$ with task-specific evaluation preferences, enhancing the model's context-aware anomaly understanding capabilities.
It serves as a verification function that checks for ideal matches between output and ground-truth answers as:
\begin{equation}
  R_\mathrm{acc}=R^{K}+\left\{\begin{array}{ll}
  R^\mathrm{TIoU}, & \text { if  }\mathcal{T}_f^V= \langle T \rangle ,\\
\lambda R^{U} + (1-\lambda)R^{H}, & \text { if  } \mathcal{T}_f^V= \langle E,\cdot \rangle , \\
R^{U}, & \text { otherwise}. \\
\end{array}\right.
\end{equation}
Here, the struct reward $R^K$, semantic reward $R^U$ and temporal reward $R^\mathrm{TIoU}$ are derived from $S^K$, $S^U$ and $S^\mathrm{TIoU}$, respectively, and $\lambda$ controls the balance between semantic and hierarchy rewards for event-related tasks.
To provide smoother hierarchy-refined guidance, we modify the hierarchy score $S^H$ by discarding the thresholding term and redefine the hierarchy reward as:
\begin{equation}
    R^H=  \frac{1}{r \cdot t} \sum_i \sum_j m_{i,j} \cdot \left(1 - \frac{d^H_{i,j}}{d^H_{\max}}\right).
\end{equation}
The overall reward used in \cue-R1 is composed of a format reward and a accuracy reward:
\begin{equation}
    R = R_\text{format} + R_\text{acc},
\end{equation}
where $R_\mathrm{format}=1$ if the response contains both $<$think$>$ and $<$answer$>$ HTML tags, otherwise $R_\mathrm{format}=0$.

Given video and prompt inputs, the policy model $\pi_\theta$ generates a group of responses containing both reasoning processes and final answers.
Each response is passed through the overall verifiable reward function ($R$) \wrt different context-aware VAU tasks to compute the reward. 
The advantage of each response ($A_i$) is then evaluated and used to update $\pi_\theta$, along with the KL-regularization from the reference model for the training stability:
\begin{align}
\mathcal{J}_\mathrm{GRPO}(\theta) 
&= \mathbb{E}_{\{\mathcal{O}_i\}_{i=1}^N \sim \pi_{\theta_\mathrm{old}}(\mathcal{O}|q)} \frac{1}{N} \sum_{i=1}^{N} \Big( \notag\\
&\quad \min \big(s\cdot A_i,\ \mathrm{clip}(s,1{-}\epsilon,1{+}\epsilon)\cdot A_i \big) \\
&\quad - \beta\, \mathbb{D}_\mathrm{KL}(\pi_\theta \parallel \pi_\mathrm{ref}) \Big), \notag
\end{align}
where $s=\frac{\pi_\theta(\mathcal{O}_i|q)}{\pi_{\theta_\text{old}}(\mathcal{O}_i|q)}$, $\epsilon$ and $\beta$ are the hyperparameters.

\begin{table*}[!ht]

\centering
\setlength{\abovecaptionskip}{5pt}
\setlength{\belowcaptionskip}{0pt}
\setlength{\tabcolsep}{1mm}
\resizebox{1\textwidth}{!}
{
\begin{tabular}{l|rrr|rr|rr|rrr|rrr|rr|rr|rrr}
\toprule
\multirow{2}{*}{\textbf{Method}} & \multicolumn{3}{c|}{\textbf{Event}} & \multicolumn{2}{c|}{\textbf{Scene}} & \multicolumn{2}{c|}{\textbf{Attribute}} & \multicolumn{3}{c|}{\textbf{Anomaly (TD)}} & \multicolumn{3}{c|}{\textbf{Anomaly (BU)}} & \multicolumn{2}{c|}{\textbf{Grounding}} & \multicolumn{2}{c|}{\textbf{Detection}} & \multicolumn{3}{c}{\textbf{Anticipation}} \\
 & \multicolumn{1}{r}{Struct} & \multicolumn{1}{r}{Sem.} & \multicolumn{1}{r|}{Hier.} & \multicolumn{1}{c}{Struct} & \multicolumn{1}{c|}{Sem.} & \multicolumn{1}{c}{Struct} & \multicolumn{1}{c|}{Sem.} & \multicolumn{1}{c}{Struct} & \multicolumn{1}{c}{Sem.} & \multicolumn{1}{c|}{Hier.} & \multicolumn{1}{c}{Struct} & \multicolumn{1}{c}{Sem.} & \multicolumn{1}{c|}{Hier.} & \multicolumn{1}{c}{Struct} & \multicolumn{1}{c|}{TIoU} & \multicolumn{1}{c}{Struct} & \multicolumn{1}{c|}{TIoU} & \multicolumn{1}{c}{Struct} & \multicolumn{1}{c}{Sem.} & \multicolumn{1}{c}{Hier.} \\
 \midrule
\rowcolor{gray!15} 
\multicolumn{21}{c}{\textit{Commercial VLMs}} \\
Gemini-1.5-flash & {60.11} & \underline{38.44} & \underline{24.23} & {84.84} &  \underline{59.90} & {29.39} &  {19.55} &  {45.08} &  \underline{39.33} &  \underline{3.11} &  \underline{57.30} &  \underline{38.57} &  \underline{3.37} &  {54.41} &  {20.74} &  {51.65} &  {21.74} &  {61.30} &  \underline{18.93} &  {\textbf{1.73}} \\
Qwen-VL-Plus &  {39.17} &  {14.40} &  {7.83} & {63.89} &  {36.92} &  {24.38} &  {4.83} &  {31.05} &  {22.91} &  {1.26} &  {27.81} & {13.77} & {0.38} &  {61.05} &  {17.44} &  {32.99} &  {7.47} & {48.17} &  {2.13} &  {0.00} \\
\midrule
\rowcolor{gray!15} 
\multicolumn{21}{c}{\textit{Open-source VLMs}} \\
Qwen2.5-VL-3B & {58.46} & {35.49} & {16.36} & {67.35} & {41.72} & {55.19} & {38.30} & \underline{53.79} & {33.80} & {1.54} & {62.66} & {30.05} & {2.11} & {44.07} & {17.73} & {63.43} & \underline{23.15} & {66.96} & {3.89} & {0.39} \\
Qwen2.5-VL-7B &  {44.36} & {19.72} &  {12.49} & {67.63} &  {41.41} &  {58.52} &  {37.72} & {16.76} &  {10.14} &  {0.73} &  {26.70} &  {16.32} &  {1.78} &  {46.66} & {17.11} & {27.27} &  {6.74} &  {80.05} &  {6.72} & {0.00} \\
InternVideo-2.5 &  {21.88} &  {12.63} &  {7.28} &  {18.76} &  {13.35} &  {9.69} & {6.08} &  {1.09} &  {1.09} & {0.11} & {29.72} &  {16.32} &  {1.17} &  {18.00} & {1.73} &  {7.93} & {1.04} &  {0.00} &  {0.00} &  {0.00} \\
Video-ChatGPT &  {22.19} & {12.78} &  {7.11} &  {17.67} &  {16.38} &  {11.39} &  {5.18} &  {1.54} &  {2.03} &  {0.14} &  {25.82} &  {14.33} &  {1.07} & {19.02} &  {1.82} &  {7.44} &  {0.89} & {0.00} &  {0.00} & {0.00} \\
Video-LLaVA & {29.33} &  {13.50} &  {8.73} &  {26.88} &  {17.21} &  {13.11} &  {9.02} &  {17.54} &  {13.75} & {0.32} & {23.24} &  {15.22} & {1.19} &  {23.04} & {3.63} &  {7.11} &  {1.15} &  {0.00} &  {0.00} &  {0.00} \\
\midrule
\rowcolor{gray!15} 
\multicolumn{21}{c}{\textit{Open-source R1 VLMs}} \\
Open-R1-Video &  {52.83} &  {30.51} &  {12.93} &  {69.08} &  {49.28} &  {48.11} &  {32.12} & {17.84} &  {13.89} &  {0.82} & {51.24} &  {21.02} & {1.79} & {32.94} & {6.24} & {4.70} & {0.85} & \underline{68.32} & {3.97} & {0.00} \\
Video-R1 & {25.23} & {9.53} &  {7.11} &  {13.99} &  {1.75} &  {47.69} &  {25.17} & {52.37} &  {35.27} &  {1.23} &  {27.22} &  {6.88} &  {0.15} &  {38.42} &  \underline{23.03} & \underline{71.81} &  {19.24} &  {27.56} &  {0.00} & {0.00} \\
Video-Chat-R1 & \underline{64.88} & {33.10} & {17.41} & \underline{86.09} & {58.03} & \underline{67.25} & \underline{45.23} & {22.86} & {14.22} &  {0.49} &  {46.93} &  {25.29} & {1.61} &  \underline{61.81} &  {20.42} &  {35.90} &  {9.27} &  {81.10} &  {11.61} & {0.00} \\
\rowcolor{green!10} 
\textbf{\cue-R1 (Ours)} &  {\textbf{83.73}} &  {\textbf{73.21}} &  {\textbf{49.16}} &  {\textbf{96.68}} &  {\textbf{82.27}} &  {\textbf{81.34}} &  {\textbf{68.14}} &  {\textbf{71.63}} &  {\textbf{67.72}} &  {\textbf{7.71}} & {\textbf{81.68}} & {\textbf{61.28}} & {\textbf{13.63}} & {\textbf{83.76}} &  {\textbf{35.94}} &  {\textbf{82.38}} &  {\textbf{35.17}} &  {\textbf{80.65}} & {\textbf{43.68}} &  \underline{0.62} \\
\bottomrule
\end{tabular}
}
\caption{\textbf{Unified Evaluation on \cuebench.} We comprehensively gauge 11 VLMs, including 10 state-of-the-art VLMs and our \cue-R1 in the unified evaluation framework. ``TD'' and ``BU'' denote top-down and bottom-up anomaly recognition, respectively. The \textbf{best} and the \underline{second-best} scores (\%) are highlighted.}
\label{tab:unified_eval}
\end{table*}

\begin{table}[!ht]
\centering
\setlength{\abovecaptionskip}{5pt}
\setlength{\tabcolsep}{1mm}
\resizebox{0.9\columnwidth}{!}
{
\begin{tabular}{cc|lc}
\toprule
\textbf{Task} & \textbf{Metric} & \textbf{Method} & \textbf{Result (\%)} \\
\midrule
\multirow{5}{*}{\begin{tabular}[c]{@{}c@{}}Event\\Recognition\end{tabular}} & \multirow{5}{*}{\begin{tabular}[c]{@{}c@{}}Top-1 / Top-5\\Hierarchy Score\end{tabular}} & CLIP  & 35.13 / 73.51 \\
 & & Open-VCLIP & 34.84 / 71.72 \\
 & & FROSTER  & 35.52 / 76.34 \\
 & & \cellcolor{gray!10}Open-MeDe  & \cellcolor{gray!10}\underline{37.03 / 76.42} \\ 
 & & \cellcolor{green!10}\textbf{\cue-R1}  & \cellcolor{green!10}\textbf{ 57.26 / 84.20 } \\
 \midrule
 \multirow{5}{*}{\begin{tabular}[c]{@{}c@{}}Temporal\\Grounding\end{tabular}} & {\multirow{5}{*}{TIoU}} & UniVTG & 17.65 \\
 & & LITA & 11.12  \\
 & & TimeChat  & 19.21 \\
 & & \cellcolor{gray!10}{UniTime}  &  \cellcolor{gray!10}\underline{21.43} \\
 & & \cellcolor{green!10}\textbf{\cue-R1}  & \cellcolor{green!10}\textbf{ 35.94 } \\
 \midrule
\multirow{8}{*}{\begin{tabular}[c]{@{}c@{}}Anomaly\\Recognition\end{tabular}} & \multirow{8}{*}{\begin{tabular}[c]{@{}c@{}}Top-1 / Top-5\\Hierarchy Score \end{tabular}} & CLIP & 10.72 / 29.89 \\
  & & Open-VCLIP  & 10.11 / 28.03 \\
 & & FROSTER & 11.97 / 33.05 \\
 & & Open-MeDe  & 12.01 / 32.83 \\
 & & VadCLIP  &  21.21 / 42.33 \\
 & & \cellcolor{gray!10}{Holmes-VAU}  &  \cellcolor{gray!10}\underline{29.72 / 53.12} \\ 
 & & \cellcolor{green!10}\textbf{\cue-R1}  & \cellcolor{green!10}\textbf{ 32.18 / 69.71 } \\
\midrule
\multirow{4}{*}{\begin{tabular}[c]{@{}c@{}}Anomaly\\Detection\end{tabular}} & \multirow{4}{*}{TIoU} & CLIP  & 13.28 \\
 & & VadCLIP  & 17.91 \\
 & & \cellcolor{gray!10}{Holmes-VAU}  & \cellcolor{gray!10}\underline{29.38}  \\
 & & \cellcolor{green!10}\textbf{\cue-R1}  & \cellcolor{green!10}\textbf{35.17} \\
\bottomrule
\end{tabular}
}
\caption{\textbf{Separate Evaluation on \cuebench.} We assess various specialized VLMs on four video understanding tasks following standard practices. }
\label{tab:separate_eval}
\vspace{-3.5mm}
\end{table}

\section{Experiment}
\subsection{Implementation Details}
We apply \cue-R1 to the Qwen2.5-VL-3B model~\cite{bai2025qwen2}, performing one epoch of supervised fine-tuning (SFT) followed by another epoch of reinforcement fine-tuning (RFT) on the \cuebench~training set, using a learning rate of $1.0e^{-6}$.
To ensure training efficiency, we cap the number of video frames at $64$, with each frame processed at a resolution of $128 \times 28 \times 28$.
For inference, we boost the frame resolution to $256 \times 28 \times 28$ and increase the number of frames to $128$ to improve performance.
Training is carried out on three NVIDIA A800 (80GB) GPUs, while inference is performed on four NVIDIA 4090 (24GB) GPUs.

\subsection{Unified Evaluation on Generative VLMs}
\tabref{unified_eval} presents a comprehensive quantitative evaluation of 10 state-of-the-art generative VLMs and our proposed \cue-R1 on \cuebench, including 2 proprietary VLMs (Gemini-1.5-Flash~\cite{team2024gemini}, Qwen-VL-Plus~\cite{bai2025qwen2}) and 8 popular open-source models, under the proposed unified evaluation framework. 
From the results, we can summarize the observations:
\textbf{1) \cue-R1 \vs Others.} Our \cue-R1 delivers a significant performance advantage across nearly all metrics of five distinct tasks, outperforming both commercial and open-source baselines. This demonstrates its effectiveness as a universal solution with strong structural alignment, high-quality semantic content and superior temporal comprehension. Notably, in complex reasoning tasks like context-aware anomaly recognition, \cue-R1 achieves semantic/hierarchy scores (\%) of $67.72/7.71$ and $61.28/13.63$ in top-down and bottom-up manners respectively, highlighting its enhanced human-aligned reasoning capabilities within event hierarchies. 
\textbf{2) Proprietary \vs Open-source VLMs.}
Compared with existing open-source VLMs, the proprietary Gemini-1.5-Flash exhibits impressive context-aware reasoning capabilities, while Qwen-VL-Plus shows marginal performance in both structural alignment and context awareness. 
\textbf{3) R1 \vs Others.}
Note that Qwen2.5-VL-3B/7B~\cite{bai2025qwen2} both achieve more promising performance across various evaluations than previous R1s.
Despite Video-R1~\cite{feng2025video} falling short in most cases, other R1-style models \ie Open-R1-Video~\cite{wang-2025-open-r1-video} and Video-Chat-R1~\cite{li2025videochat} achieve better performance than other open-source baselines like InternVideo-2.5~\cite{wang2025internvideo2}, Video-ChatGPT~\cite{maaz2023video} and Video-LLaVA~\cite{lin2023video}, highlighting their strong video reasoning capabilities. 
However, from the results, there remains considerable room for addressing the challenges of a satisfied unified solution in context-aware VAU. 

\begin{table*}[!htbp]

\centering
\setlength{\abovecaptionskip}{5pt}
    \setlength{\belowcaptionskip}{-5pt}
\setlength{\tabcolsep}{0.8mm}
\resizebox{1\textwidth}{!}
{
\begin{tabular}{l|rrr|rr|rr|rrr|rrr|rr|rr|rrr}
\toprule
\multirow{2}{*}{Method} & \multicolumn{3}{c|}{Event} & \multicolumn{2}{c|}{Scene} & \multicolumn{2}{c|}{Attribute} & \multicolumn{3}{c|}{Anomaly (TD)} & \multicolumn{3}{c|}{Anomaly (BU)} & \multicolumn{2}{c|}{Grounding} & \multicolumn{2}{c|}{Detection} & \multicolumn{3}{c}{Anticipation} \\
 & \multicolumn{1}{r}{Struct} & \multicolumn{1}{r}{Sem.} & \multicolumn{1}{r|}{Hier.} & \multicolumn{1}{c}{Struct} & \multicolumn{1}{c|}{Sem.} & \multicolumn{1}{c}{Struct} & \multicolumn{1}{c|}{Sem.} & \multicolumn{1}{c}{Struct} & \multicolumn{1}{c}{Sem.} & \multicolumn{1}{c|}{Hier.} & \multicolumn{1}{c}{Struct} & \multicolumn{1}{c}{Sem.} & \multicolumn{1}{c|}{Hier.} & \multicolumn{1}{c}{Struct} & \multicolumn{1}{c|}{TIoU} & \multicolumn{1}{c}{Struct} & \multicolumn{1}{c|}{TIoU} & \multicolumn{1}{c}{Struct} & \multicolumn{1}{c}{Sem.} & \multicolumn{1}{c}{Hier.} \\
 \midrule
Baseline & 58.5 & 35.5 & 16.4 & 67.4 & 41.4 & 55.4 & 38.3 & 53.8 & 33.8 & 1.5 & 62.7 & 30.1 & 2.1 & 44.1 & 17.7 & 63.4 & 23.2 & 67.0 & 3.9 & 0.4 \\
$+$SFT & \underline{82.4} & \underline{73.0} & \underline{46.3} & 95.9 & 81.5 & 78.9 & \underline{65.6} & 66.3 & 62.5 & \underline{7.1} & \underline{80.9} & \underline{60.8} & \underline{8.1} & 55.7 & \underline{34.6} & 51.8 & \textbf{39.0} & \underline{80.6} & \underline{39.1} & 0.0 \\
$+$RFT & 79.6 & 64.7 & 27.2 & \underline{96.6} & \underline{82.3} & \underline{80.8} & {65.0} & \underline{72.0} & \underline{67.0} & 3.1 & 80.3 & 53.8 & 2.8 & \underline{83.5} & 27.5 & \textbf{83.0} & 34.9 & 80.0 & 35.5 & \underline{0.6} \\
\rowcolor{green!10} 
\textbf{\begin{tabular}[l]{@{}l@{}}$+$SFT$+$RFT\\(Ours)\end{tabular}} & \textbf{\begin{tabular}[r]{@{}r@{}}83.7\\\textcolor{Green}{\small$\uparrow${25.2}}\end{tabular}} & \textbf{\begin{tabular}[r]{@{}r@{}}73.2\\\textcolor{Green}{\small$\uparrow${37.7}}\end{tabular}} & \textbf{\begin{tabular}[r]{@{}r@{}}49.2\\\textcolor{Green}{\small$\uparrow${32.8}}\end{tabular}} & \textbf{\begin{tabular}[r]{@{}r@{}}96.7\\\textcolor{Green}{\small$\uparrow${29.3}}\end{tabular}} & \textbf{\begin{tabular}[r]{@{}r@{}}82.3\\\textcolor{Green}{\small$\uparrow${40.9}}\end{tabular}} & \textbf{\begin{tabular}[r]{@{}r@{}}81.3\\\textcolor{Green}{\small$\uparrow${25.9}}\end{tabular}} & \textbf{\begin{tabular}[r]{@{}r@{}}68.1\\\textcolor{Green}{\small$\uparrow${29.8}}\end{tabular}} & \textbf{\begin{tabular}[r]{@{}r@{}}71.6\\\textcolor{Green}{\small$\uparrow${17.8}}\end{tabular}} & \textbf{\begin{tabular}[r]{@{}r@{}}67.7\\\textcolor{Green}{\small$\uparrow${33.9}}\end{tabular}} & \textbf{\begin{tabular}[r]{@{}r@{}}7.7\\\textcolor{Green}{\small$\uparrow${6.2}}\end{tabular}} & \textbf{\begin{tabular}[r]{@{}r@{}}81.7\\\textcolor{Green}{\small$\uparrow${19.0}}\end{tabular}} & \textbf{\begin{tabular}[r]{@{}r@{}}61.3\\\textcolor{Green}{\small$\uparrow${31.2}}\end{tabular}} & \textbf{\begin{tabular}[r]{@{}r@{}}13.6\\\textcolor{Green}{\small$\uparrow${11.5}}\end{tabular}} & \textbf{\begin{tabular}[r]{@{}r@{}}83.8\\\textcolor{Green}{\small$\uparrow${39.7}}\end{tabular}} & \textbf{\begin{tabular}[r]{@{}r@{}}35.9\\\textcolor{Green}{\small$\uparrow${18.2}}\end{tabular}} & {\begin{tabular}[r]{@{}r@{}}\underline{82.4}\\\textcolor{Green}{\small$\uparrow$\textbf{19.0}}\end{tabular}} & {\begin{tabular}[r]{@{}r@{}}
\underline{35.2}\\\textcolor{Green}{\small$\uparrow$\textbf{12.0}}\end{tabular}} & \textbf{\begin{tabular}[r]{@{}r@{}}80.7\\\textcolor{Green}{\small$\uparrow${13.7}}\end{tabular}} & \textbf{\begin{tabular}[r]{@{}r@{}}43.7\\\textcolor{Green}{\small$\uparrow${39.8}}\end{tabular}} & \textbf{\begin{tabular}[r]{@{}r@{}}0.6\\\textcolor{Green}{\small$\uparrow${0.2}}\end{tabular}} \\
\bottomrule
\end{tabular}
}
\caption{Ablation of three fine-tuning configurations based on Qwen2.5-VL-3B (Baseline). 
We maintain the same cycle length of two epochs for different training settings to ensure a fair comparison. Improvements over baseline are highlighted in \textcolor{Green}{green}.}
\label{tab:ablation}
\end{table*}

\begin{figure*}[!t]
     \setlength{\abovecaptionskip}{5pt}
    \setlength{\belowcaptionskip}{0pt}
    \centering
    \includegraphics[width=\textwidth]{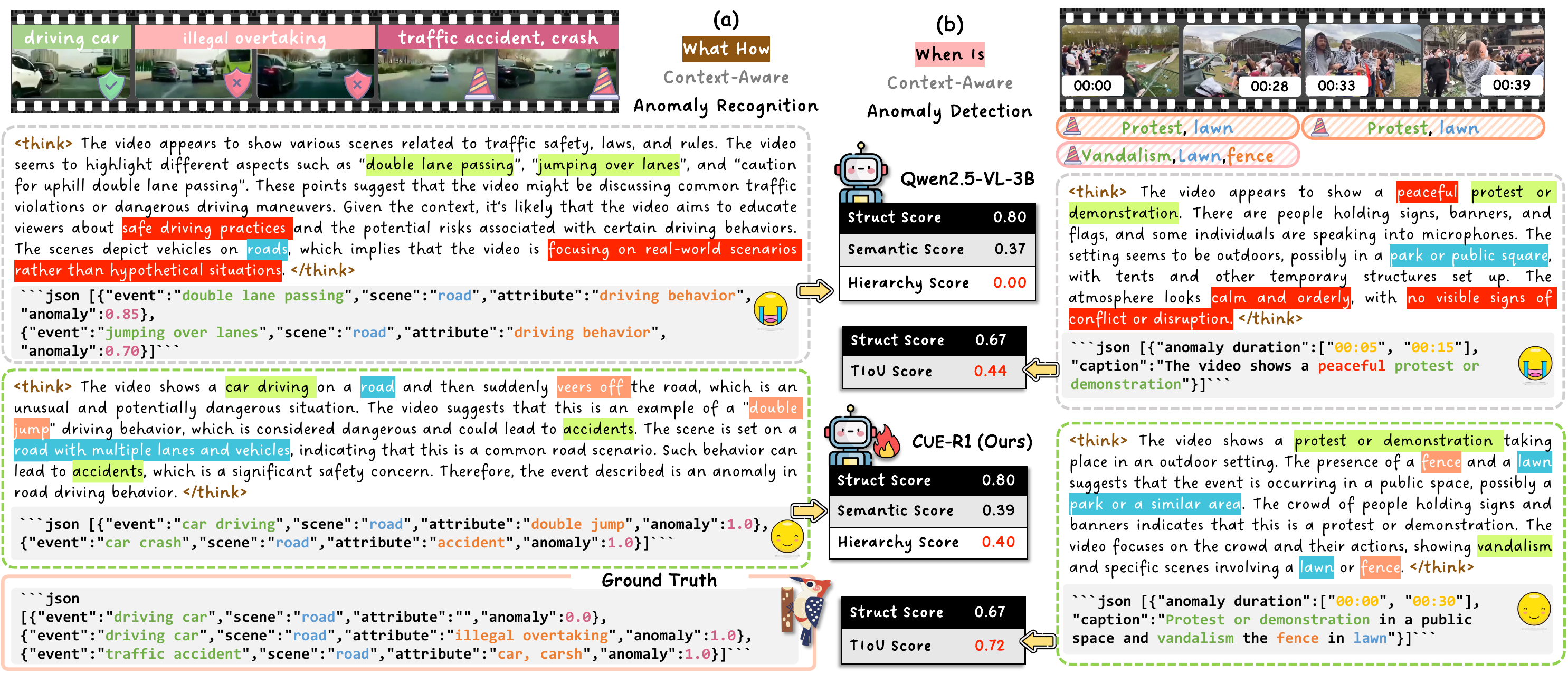}
    \caption{\textbf{Case Study.} Comparisons with Qwen2.5-VL-3B and \cue-R1 on context-aware anomaly recognition and detection.}
    \label{fig:case} 
\end{figure*}
\subsection{Separate Evaluation on Specialized VLMs}
We further conduct separate task-specific evaluations on \cuebench~following popular protocols (See Appendix) to assess various specialized VLMs on four core context-aware VAU tasks, as shown in~\tabref{separate_eval}.
Specifically, in event recognition, Open-MeDe~\cite{yu2025learning} achieves the strongest generalization among discriminative VLMs~\cite{radford2021learning,weng2023open,huang2024froster} designed for open-vocabulary action recognition.
In temporal grounding, UniTime~\cite{li2025universal} stands out as the top performer among prior methods~\cite{lin2023univtg,huang2024lita,ren2024timechat}, benefiting from elaborative training across videos of diverse contexts.
For anomaly recognition that requires context-aware capabilities, our evaluation shows that VAU methods \ie VadCLIP~\cite{wu2024vadclip} and Holmes-VAU~\cite{zhang2025holmes} significantly outperform general action recognition approaches.
Holmes-VAU records strong performance for both anomaly recognition and detection, indicating its superior anomaly understanding capabilities.
Overall, \cue-R1 outperforms both discriminative and generative specialized VLMs across tasks. 
Despite the strengths of task-specific models, their performance still manifests clear limitations, particularly in semantic alignment and anomaly reasoning, underscoring the advantage of our unified and context-aware generative approach.

\subsection{Ablation Study}
To assess the contributions of SFT and RFT strategies, we conduct an ablation study by performing two variants on Qwen2.5-VL-3B model. 
The results in~\tabref{ablation} clearly demonstrate the effectiveness of both strategies in our training pipeline.
Compared to the baseline, SFT yields substantial gains especially in semantic scores, demonstrating its effectiveness in enhancing alignment with structured answers and improving content consistency.
While RFT alone brings more gains over SFT on struct scores, the hierarchy scores improve only marginally or stagnate, suggesting that reward signals based solely on task performance could be insufficient to capture fine-grained semantic relations or hierarchical distinctions.
By combining SFT and RFT sequentially, \cue-R1 achieves the best overall performance, serving as a robust and context-aware generative VLM for comprehensive VAU. 
The large improvement in hierarchy scores, especially for complex tasks like anomaly recognition, validates the benefit of incorporating human-aligned hierarchical feedback in RFT.
The comparison highlights that SFT provides strong structural and semantic grounding, while RFT complements it by refining task alignment. 

\subsection{Case Study}
\figref{case} presents qualitative and quantitative comparisons between Qwen2.5-VL-3B and \cue-R1 on two representative VAU tasks under the unified evaluation paradigm.
(a) For anomaly recognition, Qwen2.5-VL-3B fails to recognize the severity and specificity of the anomalies. It correctly identifies the scene and mentions \textit{jumping over lanes}, yet misrepresenting dangerous maneuvers as generic traffic behavior.
\cue-R1, in contrast, identifies two well-grounded events: \textit{car driving} and \textit{car crash}, associating them with meaningful attributes like \textit{double jump} and \textit{accident}. This reflects an accurate contextual and semantic interpretation of the anomaly. It scores higher on hierarchy metrics, reflecting better alignment within the event hierarchy.
(b) For anomaly detection, Qwen2.5-VL-3B offers surface-level reasoning process: ``\textit{a peaceful protest or demonstration}'', lacking the anomaly relevant details (e.g., vandalism) and struggles with hallucination and poor anomaly localization.
Conversely, \cue-R1 delivers contextually grounded, semantically rich, and temporally precise predictions, highlighting its superior performance for anomaly understanding.

\section{Conclusion}
This paper presents \cuebench, the first large-scale benchmark for evaluating the context-aware video anomaly understanding capabilities of VLMs in a unified framework. 
We establish a comprehensive event-centric hierarchical taxonomy with absolute and conditional anomaly events and diverse context-aware anomalies and normalities.
Our extensive evaluation highlights significant performance gaps remaining among existing state-of-the-art VLMs.
Building upon this, we propose \cue-R1, an R1-style method that outperforms the leading VLMs by a notable margin on \cuebench. 
This work not only provides a solid foundation for developing unified generative VLMs, but also serves as a challenging benchmark for VAU in real-world.

{\small
\bibliography{aaai2026}
}
\clearpage

\appendix

\section*{Supplementary Material}
This supplementary material offers extensive additional details, experimental results and discussions complementing the main paper. The content is organized as follows:
\begin{enumerate}[label=\Alph*.]
    \item Details of Dataset (\appenref{dataset})
    \item Details of \cue-R1 (\appenref{cue-r1})
    \item Details of Unified Evaluation (\appenref{evaluation})
    \item Additional Implementation Details (\appenref{implementation})
    \item Additional Experimental Results (\appenref{experiments})
    \item Discussions (\appenref{discussions})
\end{enumerate}

\section{Details of Dataset}
\label{appen:dataset}

\subsection{Dataset Engineering}
\label{appen:engineering}
\paragraph{Dataset Collection.}
To address the limitations of existing datasets, we introduce a new VAU dataset comprising untrimmed web videos sourced from YouTube.
For a broad coverage of the dataset, we first curated over 7,000 YouTube videos depicting real-world absolute and conditional anomaly events. 
As the web videos present extreme diversity and complexity, spanning genres, camera views, editing and compositing, we apply a rigorous filtering process to ensure the quality, ethics and appropriateness of the dataset. 
The final dataset consists of 2,950 videos (at a fixed rate of 30 fps) that are thematically coherent, contextually rich, and focused on real-world anomalies.

\begin{figure}[hp]
    \centering
    \includegraphics[width=0.4\columnwidth]{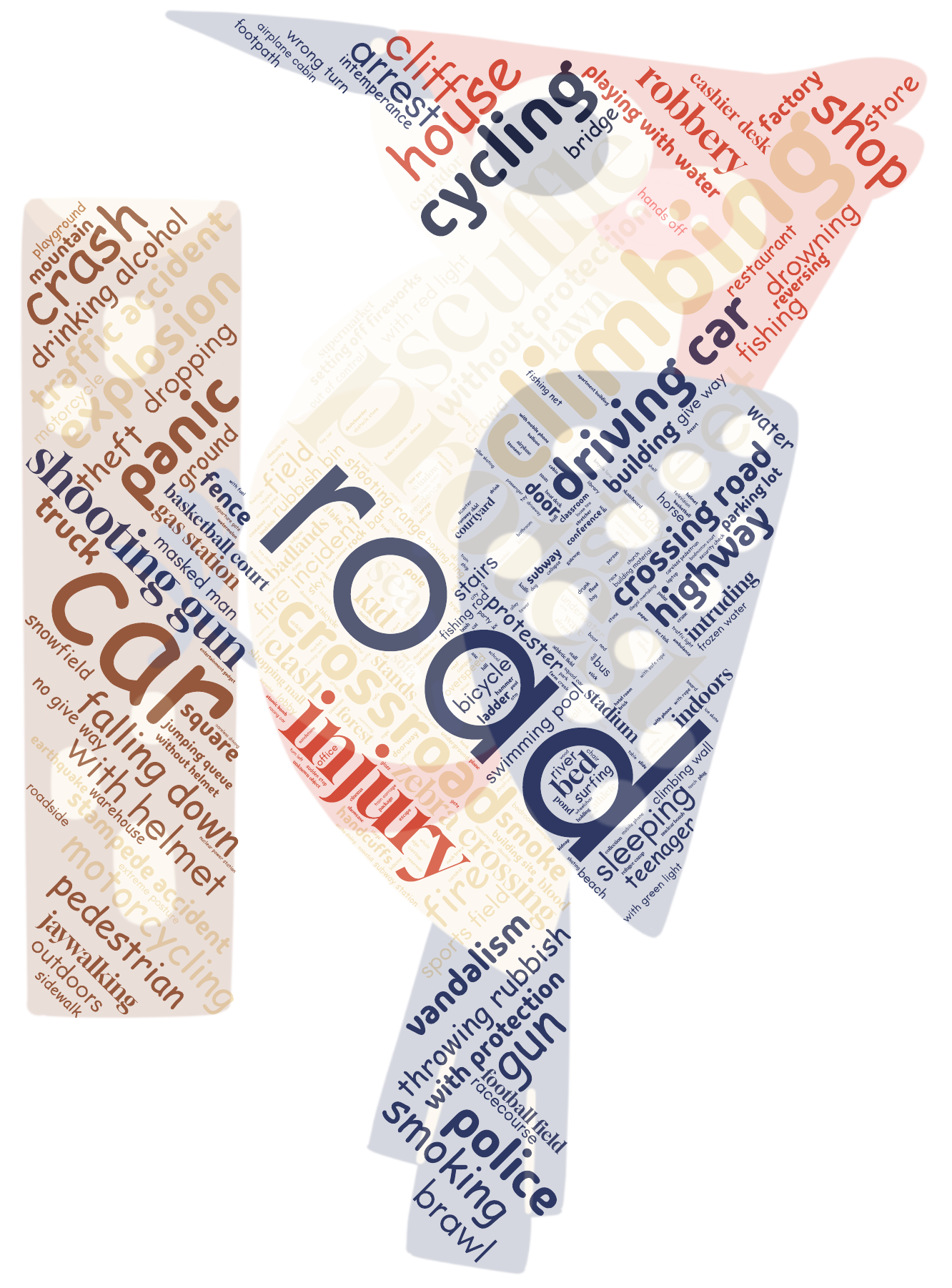}
    \caption{Word cloud of the contexts in \cuebench.}
    \label{fig:wordcloud} 
\end{figure}

\paragraph{Manual Annotation.}
Due to potential noise in raw data and the need for fine-grained labeling, our annotation pipeline follows a three-stage process: context-aware anomaly annotation (instance-level), anomaly localization (frame-level), and content integrity review.
\begin{itemize}
    \item \textit{Context-aware Anomaly Annotation.}
    In this stage, we primarily annotate the context triplets, denoted as $C=\left \langle E,S,A \right \rangle $, \ie the co-occurrences of events ($E$), scenes ($S$) and attributes ($A$) within each video.
    Additional metadata, such as video genres and camera views, are also recorded.
    The word cloud in~\figref{wordcloud} illustrates the distribution of annotated contexts in our dataset.
    Each triplet is assigned a binary anomaly label: if the triplet suggests an abnormal occurrence, it is labeled as anomalous ($N_{\left \langle E,S,A \right \rangle}=1$); otherwise, it is marked as normal ($N_{\left \langle E,S,A \right \rangle}=0$).

    \item \textit{Context-aware Anomaly Localization.}
    For each annotated triplet, we temporally localize its occurrence by marking the precise start and end frames. This step effectively constitutes a form of anomaly localization, given that each triplet has already been classified as anomalous or normal.

    \item \textit{Content Integrality Review.} 
    Notably, the annotation process is highly challenging due to fine-grained contexts, complex anomaly dependencies, and ambiguous temporal boundaries.
    To ensure the quality, we rigorously verify instance-level annotations for the context coherence and integrality. 
    Specifically, we cross-check that the scene ($S$) and attribute ($A$) annotations sufficiently differentiate anomalies from normalities sharing the same event ($E$), and vice versa.
    Additionally, we validate frame-level annotations for temporal consistency across different annotators, accounting for variations in time formats (\eg frame indices \vs timestamps in seconds). 
\end{itemize}

\begin{figure*}[!htpb]
    \centering
    \includegraphics[width=\textwidth]{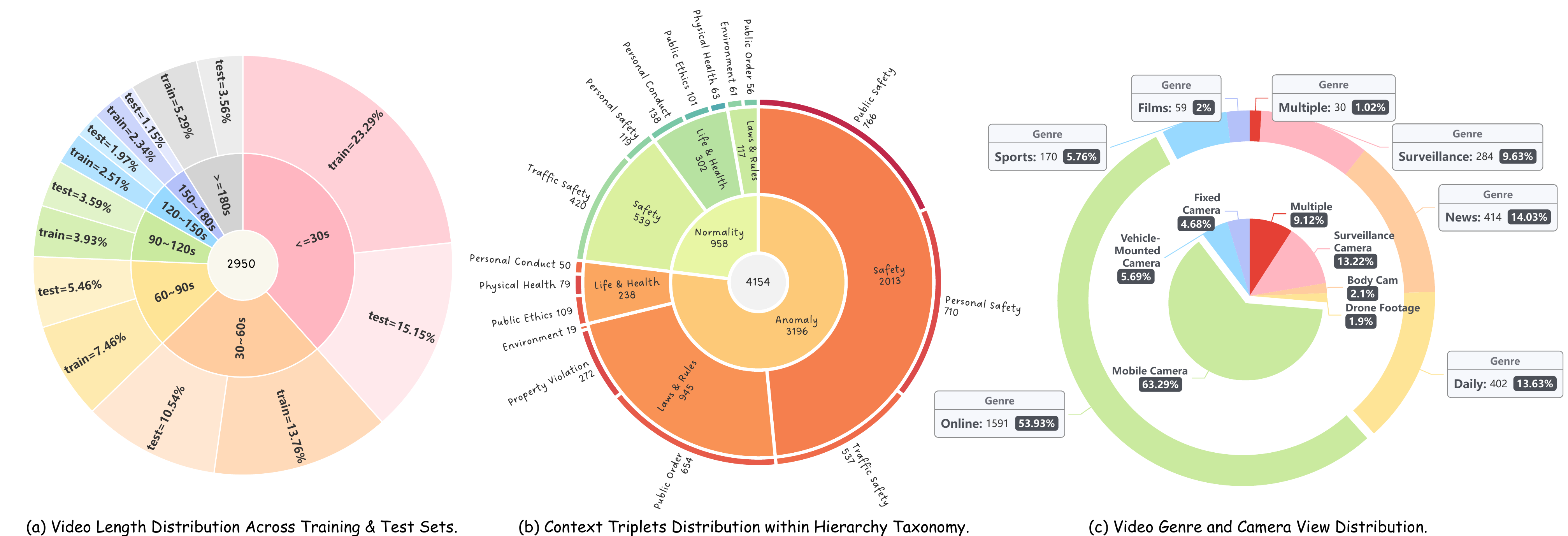}
    \caption{{Additional statistical analysis of \cuebench.} }
    \label{fig:A_statistics} 
\end{figure*}

\begin{figure*}[!htpb]
    \centering
    \begin{subfigure}[t]{\columnwidth}
        \centering
        \includegraphics[width=\columnwidth]{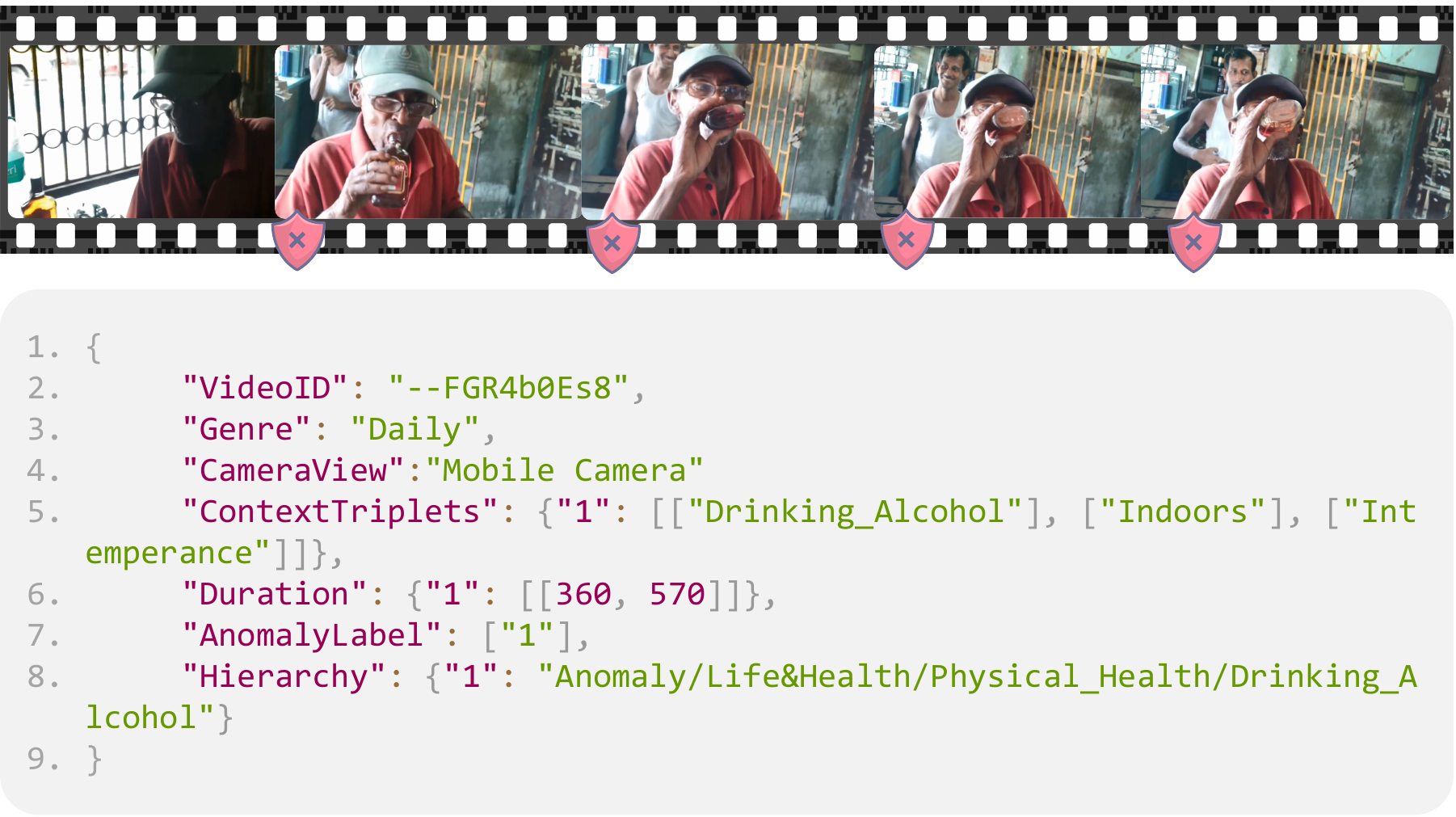}
        \caption{Conditional anomaly for ``Drinking Alcohol''.}
        \label{fig:eg1}
    \end{subfigure}
    \hfill
    \begin{subfigure}[t]{\columnwidth}
        \centering
        \includegraphics[width=\columnwidth]{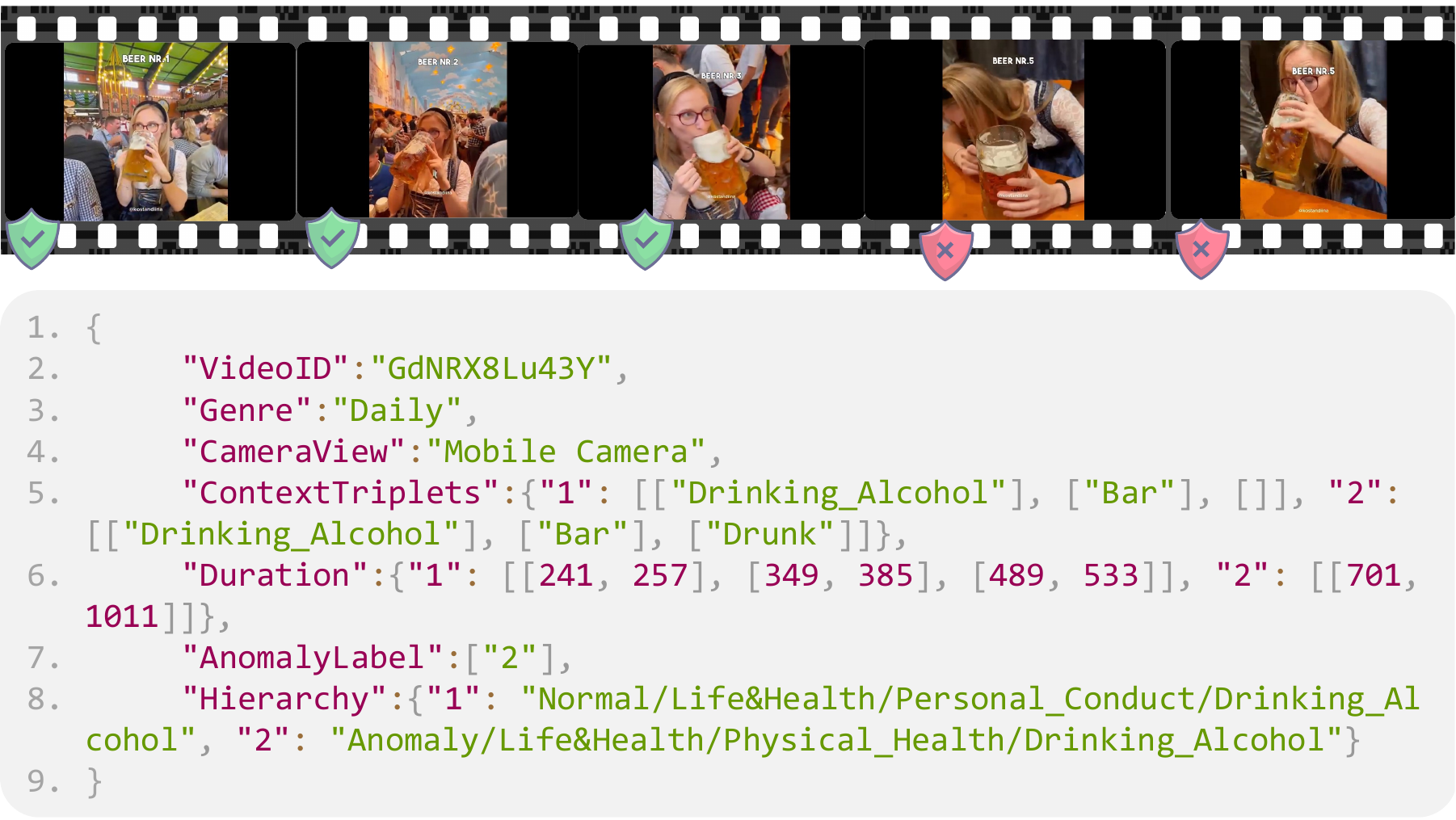}
        \caption{Conditional anomaly and normality for ``Drinking Alcohol''.}
        \label{fig:eg2}
    \end{subfigure}
    \vskip\baselineskip
    \begin{subfigure}[t]{\columnwidth}
        \centering
        \includegraphics[width=\columnwidth]{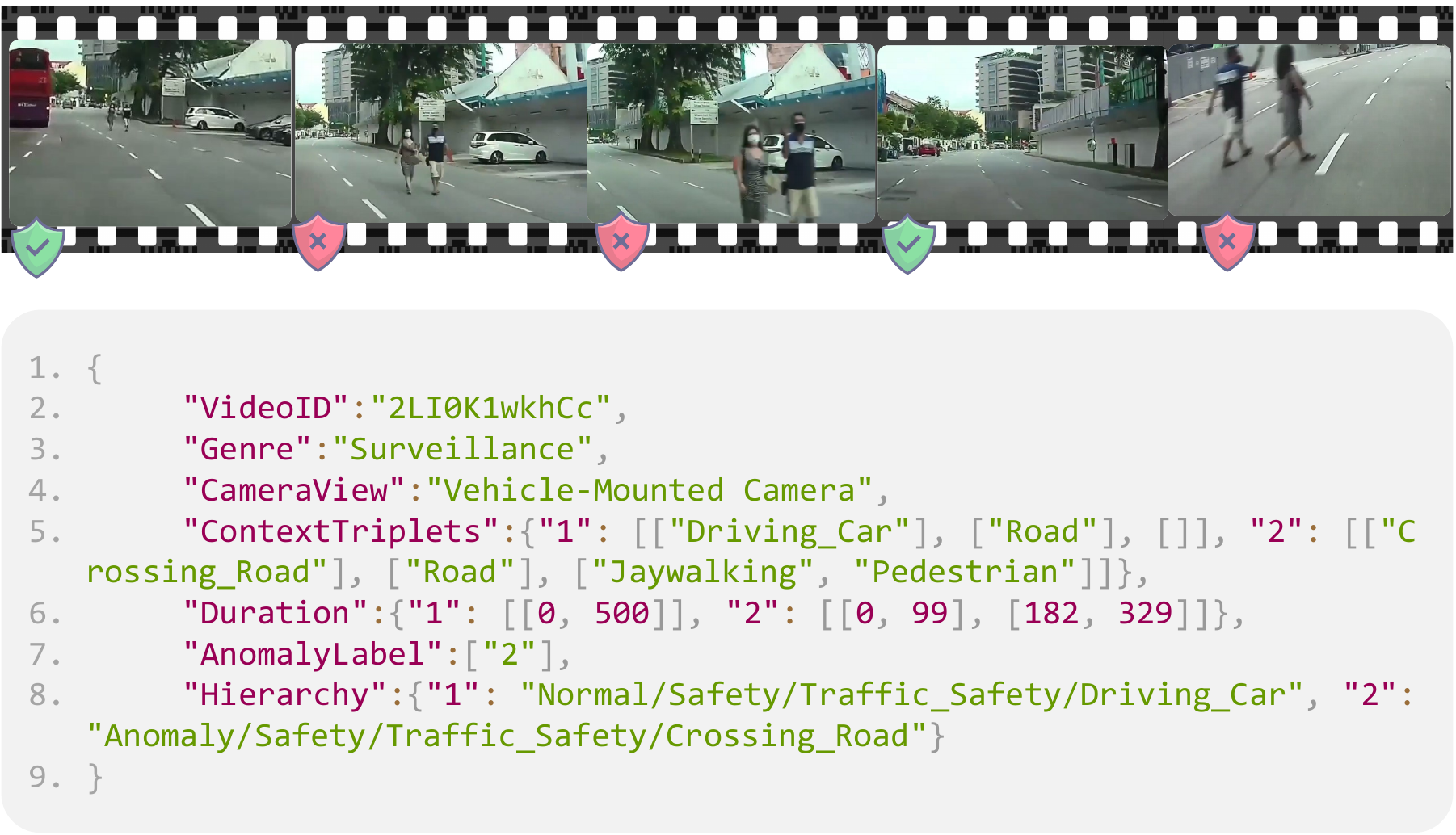}
        \caption{Conditional normality for ``Driving Car'' and conditional anomaly for ``Crossing Road''.}
        \label{fig:eg3}
    \end{subfigure}
    \hfill
    \begin{subfigure}[t]{\columnwidth}
        \centering
        \includegraphics[width=\columnwidth]{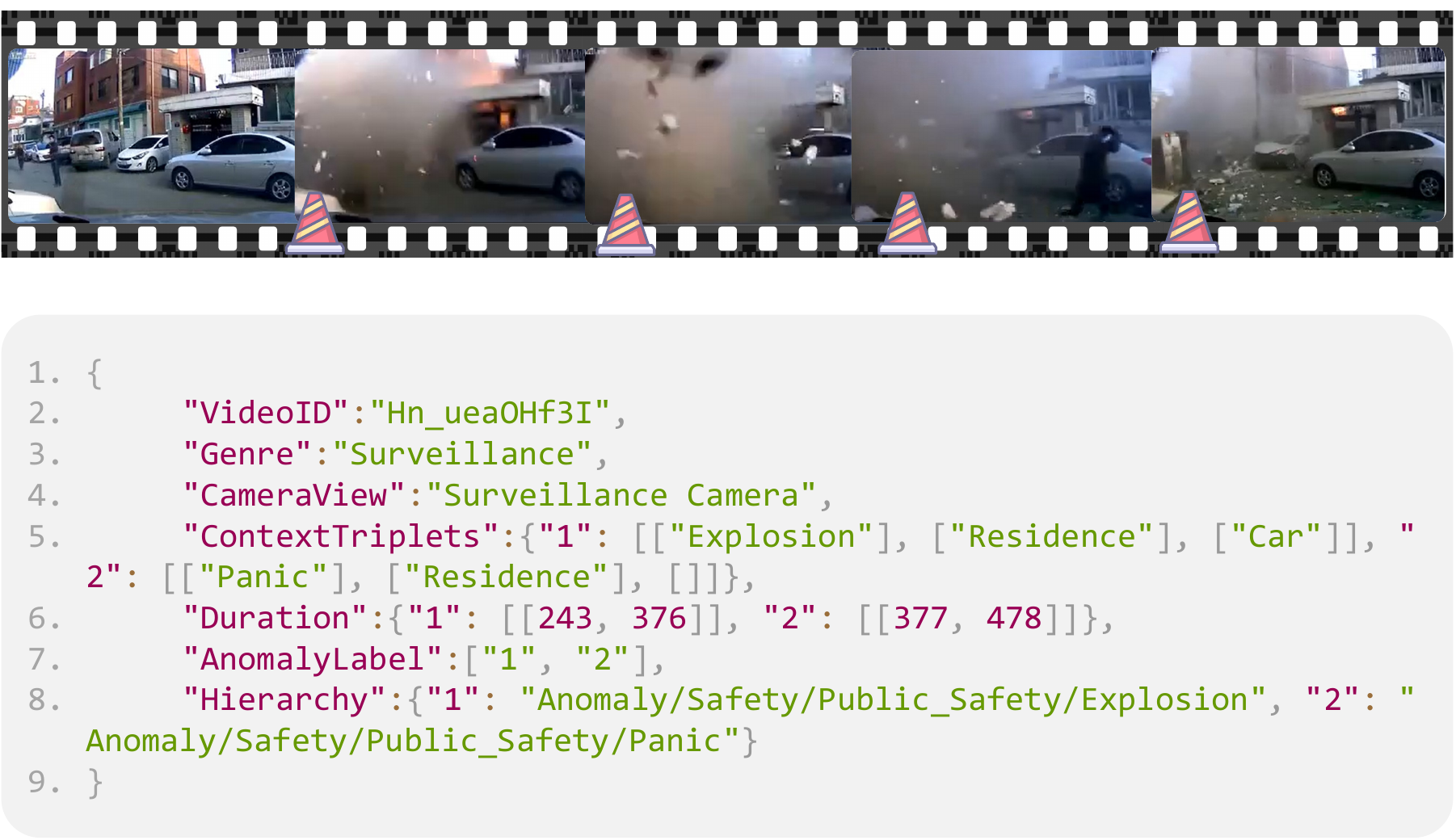}
        \caption{Absolute anomaly for ``Explosion'' and ``Panic''.}
        \label{fig:eg4}
    \end{subfigure}
    \caption{Annotation examples in \cuebench.}
    \label{fig:egs}
\end{figure*}

\subsection{More Data Statistics}
\label{appen:statistics}
\figref{A_statistics} provides additional statistics for the proposed \cuebench.
The overall video length distribution for the training and test data is shown in~\figref{A_statistics}(a).
Our dataset emphasizes a substantial number of short videos (under 60 seconds) and longer videos, ensuring exposure to extended temporal reasoning, making the dataset versatile for a wide range of video understanding challenges. 
The relatively balanced distribution between training and test data across different durations supports fair performance evaluation and robust generalization.
On average, each video lasts 66.5 seconds and contains 1.4 context triplets, which span approximately 62\% of the total video duration.
\figref{A_statistics}(b) shows the distribution of context triplets within the hierarchical taxonomy.
For a total occurrence of 4,154 context triplets, the anomalies constitute the majority of the dataset, with the safety domain particularly prominent, comprising over half of the anomaly examples, suggesting strong coverage of personal, public, and traffic domains.
The normality branch also mirrors the rich context triplets from the comprehensive hierarchical taxonomy.
The average duration for each anomaly and normality in our dataset is 25.3s and 42.1s, respectively.
This analysis highlights that \cuebench~captures a broad spectrum of rapid anomalies and sustained normalities, enabling a comprehensive evaluation of context-aware VAU in real-world.
Finally, the video proportion by genre and camera view in our \cuebench~is presented in~\figref{A_statistics}(c).

\subsection{Video Annotation Examples}

As shown in~\figref{egs}, we provide the final constructed annotation examples in JSON format, which include:
\begin{itemize}
    \item Video IDs: \textit{String} (from YouTube)
    \item Genres: \textit{Type}
    \item Camera Views: \textit{Type}
    \item Context Triplets: \textit{Dictionary}
    \item Frame Durations: \textit{Dictionary}
    \item Anomaly Labels: \textit{List}
    \item Hierarchical Taxonomy Categorizations: \textit{Dictionary}
\end{itemize}

\begin{figure*}[!t]
    \centering
    \includegraphics[width=0.9\textwidth]{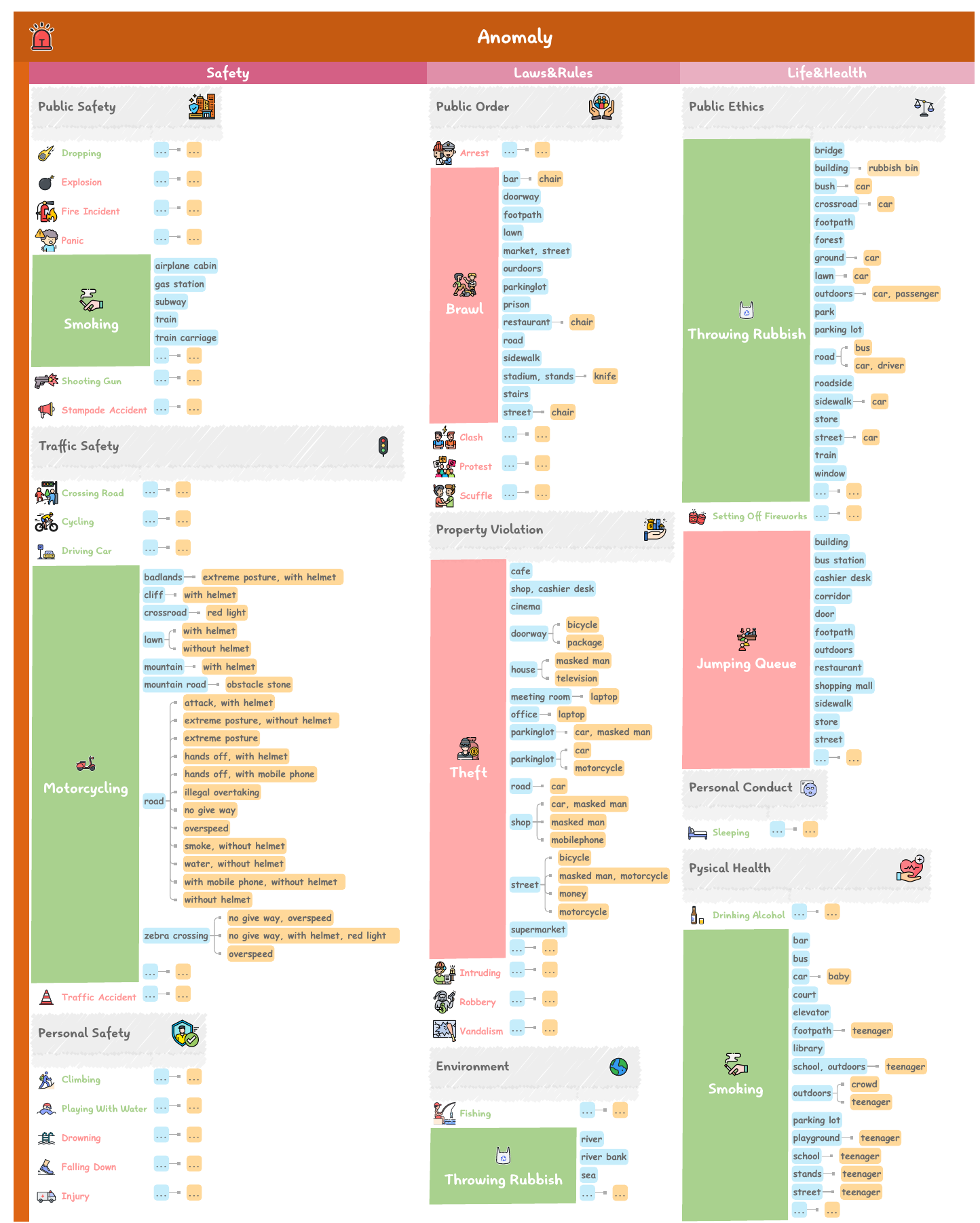}
    \caption{{Detailed abnormal context triples within hierarchy taxonomy for absolute and conditional anomaly events.} }
    \label{fig:anomaly} 
\end{figure*}
\begin{figure*}[t]
    \centering
    \includegraphics[width=\textwidth]{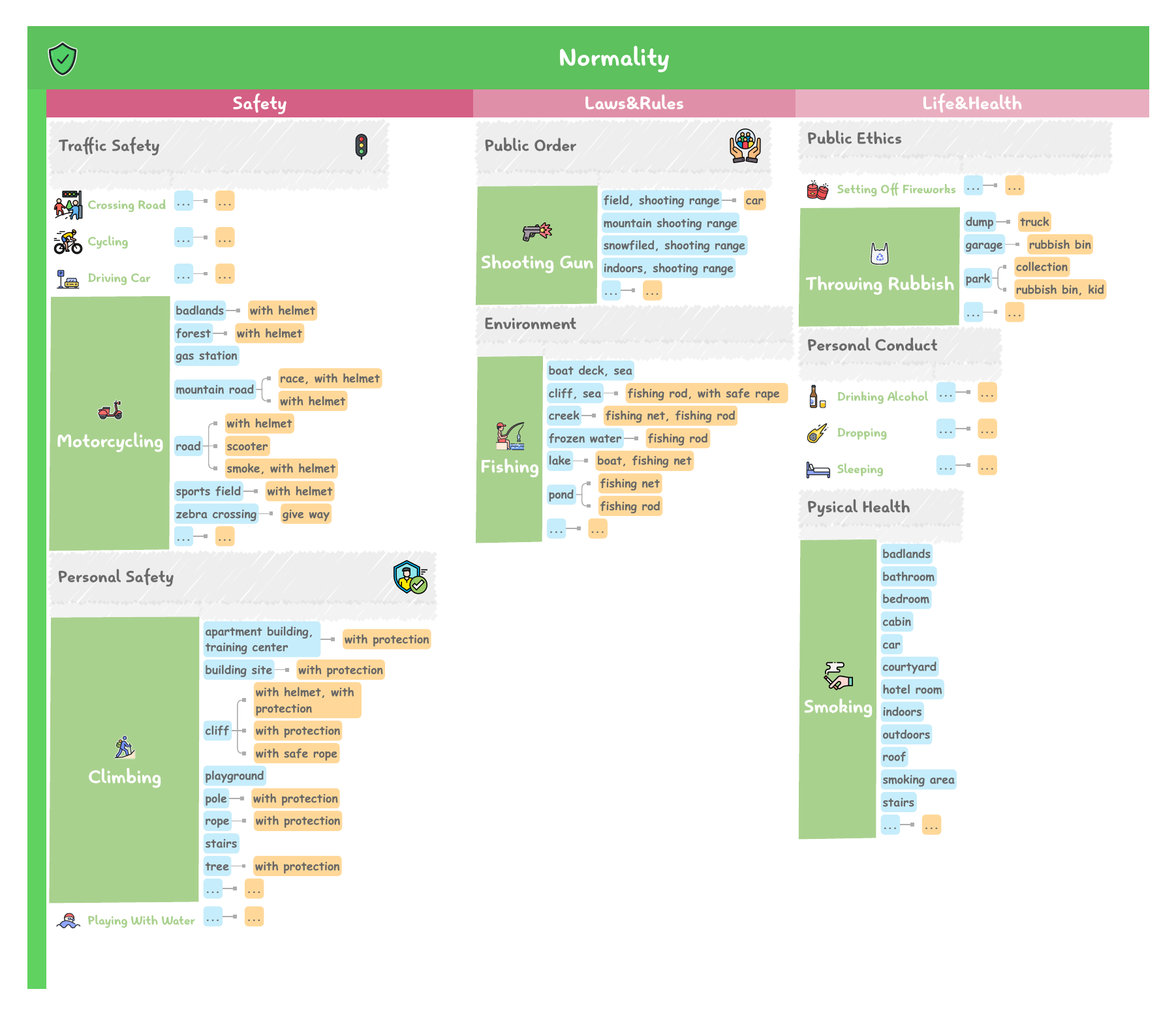}
    \caption{\textbf{Detailed normal context triples within hierarchy taxonomy for conditional anomaly events.} }
    \label{fig:normality} 
\end{figure*}
\subsection{Hierarchy Taxonomy Details}
\label{appen:hierarchy}

\figref{anomaly} shows the anomaly branch within the constructed hierarchy taxonomy. We provide the detailed abnormal context triplets for three absolute anomaly events (\textit{Brawl}, \textit{Theft} and \textit{Jumping Queue}) and three conditional anomaly events (\textit{Smoking}, \textit{Motorcycling}, and \textit{Throwing Rubbish}) covering various domains and effects.

\figref{normality} shows the normality branch of the hierarchy taxonomy. We provide the detailed normal context triplets of six conditional anomaly events from various domains and effects, \ie \textit{Motorcycling}, \textit{Climbing}, \textit{Shooting Gun}, \textit{Fishing}, \textit{Throwing Rubbish} and \textit{Smoking}.

In the domain of safety, the focus is specifically on behavioral and physical violations (\eg \textit{motorcycling without a helmet}), while the characteristics lean toward criminal intent or social disruption for the domain of laws \& rules.
For life \& health, anomaly judgments are more ``normative'', relying on social roles (\eg \textit{teenager drinking}) or environmental sensitivity (\eg \textit{throwing rubbish in a river}).
This hierarchical and contextual organization allows comprehensive coverage, supporting nuanced interpretation of both anomalies and normalities.

\textbf{\textit{Scenes play distinct semantic roles in the effects.}}
Note that identical events can lead to qualitatively different outcomes, depending on ``where'' and ``how'' the event unfolds. 
This emphasizes that anomaly understanding is not only about identifying ``what is unusual'', but also ``why it matters'' \ie its implications vary by contexts.
For example, \textit{smoking on a bus} is categorized under life \& health, primarily due to its adverse effects on physical health of nearby passengers in a confined space.
In contrast, \textit{smoking at a gas station} falls under public safety, as it poses a serious fire risk and potential for explosion, with consequences far beyond individual health.

\textit{\textbf{Attributes vary not just in presence but in semantic types.}}
In \cuebench, attributes serve as essential context, enriching anomaly detection.
For absolute anomaly events, which are inherently more inclined toward being abnormal, attributes play a less decisive role. 
Conversely, for conditional anomalies, attributes are crucial for distinguishing normal from abnormal, making them both more critical and extensively annotated.
Specifically, physical attributes include equipment (\eg helmet, bicycle), age (\eg teenager, baby), or posture (\eg extreme posture); social attributes reflect roles and intentions (\eg masked man, crowd, car driver), enriching interpretation through implied risks or goals; behavioral attributes describe action nuances (\eg illegal overtaking, failure to yield way, use of mobile phone), enabling subtle abnormality distinctions even within the same event.
Our detailed construction requires models to interpret anomalies not merely by labels, but through compositional reasoning based on diverse contextual cues.
This rich variety underscores the need for context-aware methods capable of both low-level perception and high-level reasoning.
In effect, our context triplets are highly expressive, capturing fine-grained cues essential for understanding both absolute and conditional anomaly events.



\section{Details of \cue-R1}
\label{appen:cue-r1}

\subsection{GRPO in \cue-R1}
\label{appen:GRPO}
\begin{figure*}[!htbp]
    \centering
    \includegraphics[width=\textwidth]{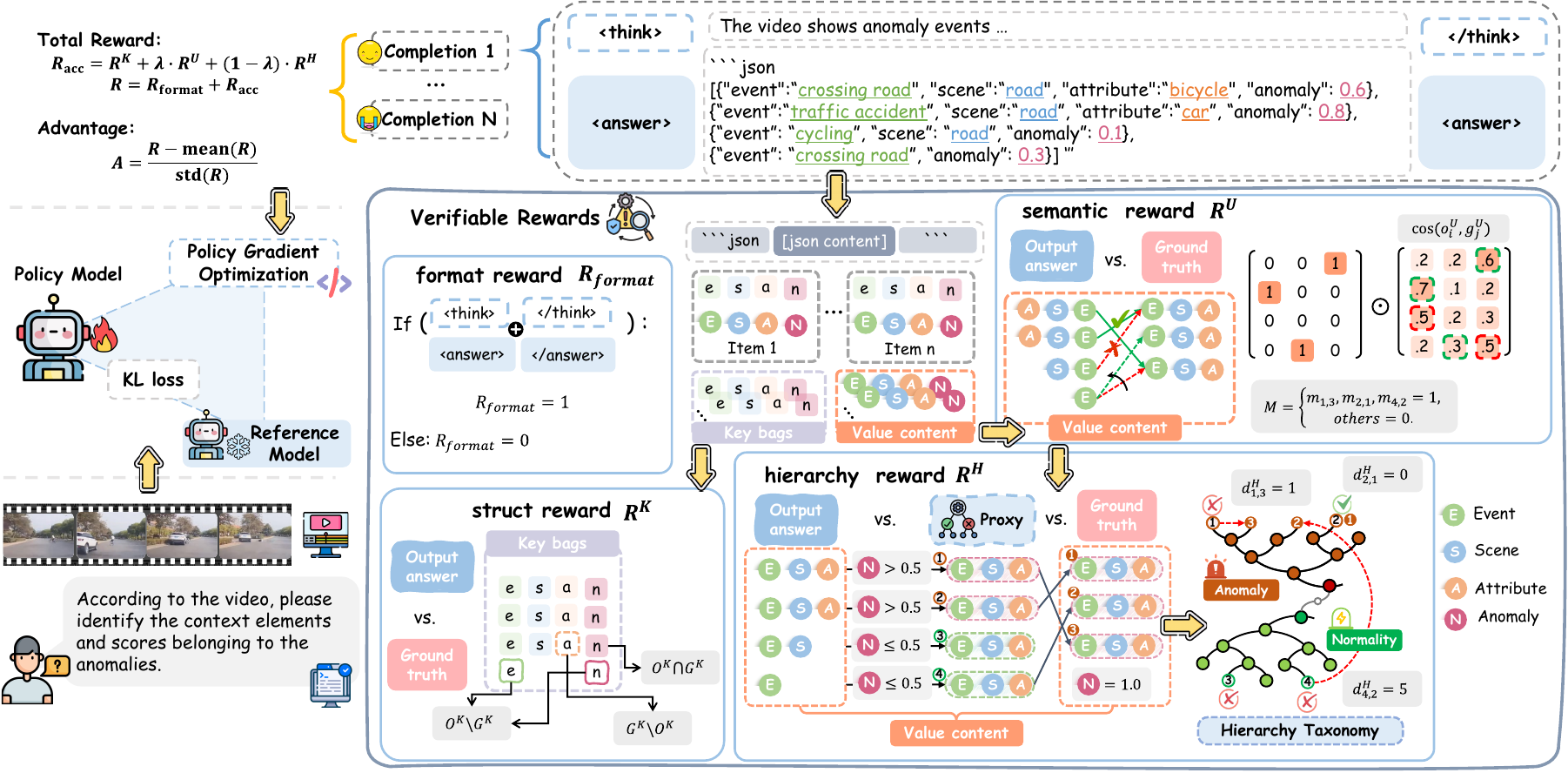}
    \caption{\textbf{\cue-R1's GRPO algorithm.} Given the prompts and video inputs, the policy model generates multiple completions. Then the verifiable reward of the format and accuracy is used with the policy gradient optimization algorithm to update the policy model.}
    \label{fig:cue-r1} 
\end{figure*}

Unlike reinforcement learning algorithms such as PPO~\cite{schulman2017proximal}, which require an additional critic model to estimate policy performance, we adopt the GRPO algorithm~\cite{guo2025deepseek}, which directly compares a set of candidate completions without relying on a separate critic.
We present the detailed verifiable accuracy rewards of \cue-R1 in~\algref{alg:grpo}.
The policy model is prompted to generate $N$ candidate completions, each containing both a reasoning process and a final answer structured as a list of key-value pairs, \ie in JSON format.
GRPO then evaluates each completion using our custom-designed verifiable format and accuracy rewards, which assess both structural integrity and content quality based on the given task.
Specifically, output answers are first parsed using a JSON parser. 
The extracted keys are assessed using the structural reward $R^K$. 
The values are then compared with the ground truth using task-aligned rewards: the temporal IoU reward $R^U$ for ``When'' tasks, the semantic reward $R^K$ for ``What'' tasks, and the hierarchy reward $R^H$ for event-related ``What'' tasks.
\figref{cue-r1} illustrates the GRPO algorithm applied in \cue-R1 for bottom-up, context-aware anomaly recognition. 
The semantic reward is calculated based on cosine similarity between the triplet representations of the predicted answers and ground-truths, using a binary matching matrix to prevent reward hacking.
For hierarchy evaluation, if the predicted anomaly score $>0.5$, we retrieve proxy triplets from the anomaly hierarchy; otherwise, we use the normality hierarchy. 
The hierarchy reward is then calculated based on both the semantic matching matrix and the hierarchy distance between the proxies and ground-truths to ensure hierarchy-aware verification.

\subsection{Training Process of \cue-R1}
\label{appen:train}

\algref{alg:training} shows the training process of our proposed \cue-R1.
Both supervised fine-tuning (SFT) and reinforcement fine-tuning (RFT) are conducted on the same dataset, which includes videos, task-specific prompts, format instructions, and ground-truth annotations across all five tasks.
We begin with SFT using auto-regressive cross-entropy loss to teach the model to follow instructions.
Following this, we apply RFT with GRPO to further optimize the policy model.
The SFT phase equips the model with foundational instruction-following abilities across tasks, while the RFT phase moves beyond rigid pattern matching, encouraging the model to explore and adopt more flexible, context-aware reasoning strategies for anomaly understanding.

\begin{algorithm}[!htbp]
\caption{\textbf{Verifiable accuracy reward of \cue-R1}}
\label{alg:grpo}
\small
\textbf{Input}: Video $\mathcal{V}$, problem prompt $\mathcal{T}_p$ and format prompt $\mathcal{T}_f=\{\mathcal{T}_f^K,\mathcal{T}^V_f\}$ of task $\mathcal{T}$ \\
\textbf{Require}: Policy model $\pi_\theta$, ground-truth $\mathcal{G}=\{\mathcal{G}^K,\mathcal{G}^V\}$, hyperparameter $\lambda$. \\
\textbf{Output}: Accuracy reward $R_\text{acc}$.
\begin{algorithmic}[1]
\STATE Generate $N$ completions: $\{\mathcal{O}_i, \mathcal{R}_i\}_{i=1}^N\gets \pi(\mathcal{V},\mathcal{T}_p,\mathcal{T}_f)$
\STATE Init accuracy reward: $R_\text{acc}=\{r_i\}_{i=1}^N$, where $r_i=0$
\FOR{each $\mathcal{O}_i$} 
\STATE Extract key bags $\mathcal{O}^K_i$ and value content $\mathcal{O}^V_i$ from completions $\mathcal{O}_i$ \wrt $\mathcal{T}_f^K$ and $\mathcal{T}_f^V$
\STATE Calculate structure reward: $R^K=S^K_{\mathcal{T}_f^K}(\mathcal{O}_i^K,\mathcal{G}^K)$
\IF {$\mathcal{T}_f^V=\langle T \rangle$}
\STATE Compute temporal reward: $R^{\mathrm{TIoU}}=S^{\mathrm{TIoU}}_{\langle T \rangle}(\mathcal{O}_i^V,\mathcal{G}^V)$
\STATE Obtain $r_i\gets R^K+R^{\mathrm{TIoU}}$
\ELSE
\STATE Compute semantic reward: $R^U=S^U_{\mathcal{T}_f^V}(\mathcal{O}_i^V,\mathcal{G}^V)$
\ENDIF
\IF {$\mathcal{T}_f^V=\langle E, \rangle$}
\STATE Compute hierarchy reward: $R^H=S^H_{\mathcal{T}_f^V}(\mathcal{O}_i^V,\mathcal{G}^V;\tau)$, where $\tau=1$
\STATE Obtain $r_i\gets R^K+\lambda R^U+(1-\lambda)R^H$
\ELSE 
\STATE Obtain $r_i\gets R^K+ R^U$
\ENDIF
\ENDFOR
\STATE \textbf{return} $R_\text{acc}=\{r_i\}_{i=1}^N$
\end{algorithmic}
\end{algorithm}

\begin{algorithm}[!htbp]
\caption{\textbf{\cue-R1 training process}}
\label{alg:training}
\small
\textbf{Input}: Training set $\mathbb{D}=\{(\mathcal{V},\mathcal{T}_p,\mathcal{T}_f,\mathcal{G})\}^M$. \\
\textbf{Require}: Policy model $\pi_{\theta_\text{init}}$. \\
\textbf{Output}: Final policy model $\pi_\theta$.
\begin{algorithmic}[1]
\STATE Init policy model: $\pi_\theta \gets \pi_{\theta_\text{init}}$ \\
 \texttt{// perform supervised fine-tuning}
\FOR {each $(\mathcal{V},\mathcal{T}_p,\mathcal{T}_f,\mathcal{G})\in\mathbb{D}$} 
\STATE Calculate cross-entropy loss: $\mathcal{L}_{s}(\theta)=-\log p_\theta(\mathcal{G}|\mathcal{V},\mathcal{T}_p)$
\STATE Update $\pi_\theta$ \wrt $\mathcal{L}_{s}(\theta)$
\ENDFOR\\
\texttt{// perform reinforcement fine-tuning}
\STATE Init reference model: $\pi_{\text{ref}}\gets\pi_\theta$ 
\FOR {each $(\mathcal{V},\mathcal{T}_p,\mathcal{T}_f,\mathcal{G})\in\mathbb{D}$} 
\STATE Generate $N$ completions: $\{\mathcal{O}_i, \mathcal{R}_i\}_{i=1}^N\gets \pi(\mathcal{V},\mathcal{T}_p,\mathcal{T}_f)$
\STATE Compute format reward: $R_{\text{format}}$
\STATE Compute accuracy reward: $R_{\text{acc}} \gets$ \algref{alg:grpo}
\STATE Calculate total reward: $R=R_{\text{format}}+R_{\text{acc}}$
\STATE Calculate advantages: $A=\frac{R-\text{mean}(R)}{\text{std}(R)}$
\STATE Update $\pi_\theta$ \wrt $\mathcal{J}_{\text{GRPO}}(\theta)$
\ENDFOR
\RETURN $\pi_\theta$
\end{algorithmic}
\end{algorithm}

\section{Details of Unified Evaluation}
\label{appen:evaluation}
\subsection{Prompts of Unified Evaluation}
\label{appen:prompt}
\begin{figure*}[ht]
    \centering
    \includegraphics[width=\textwidth]{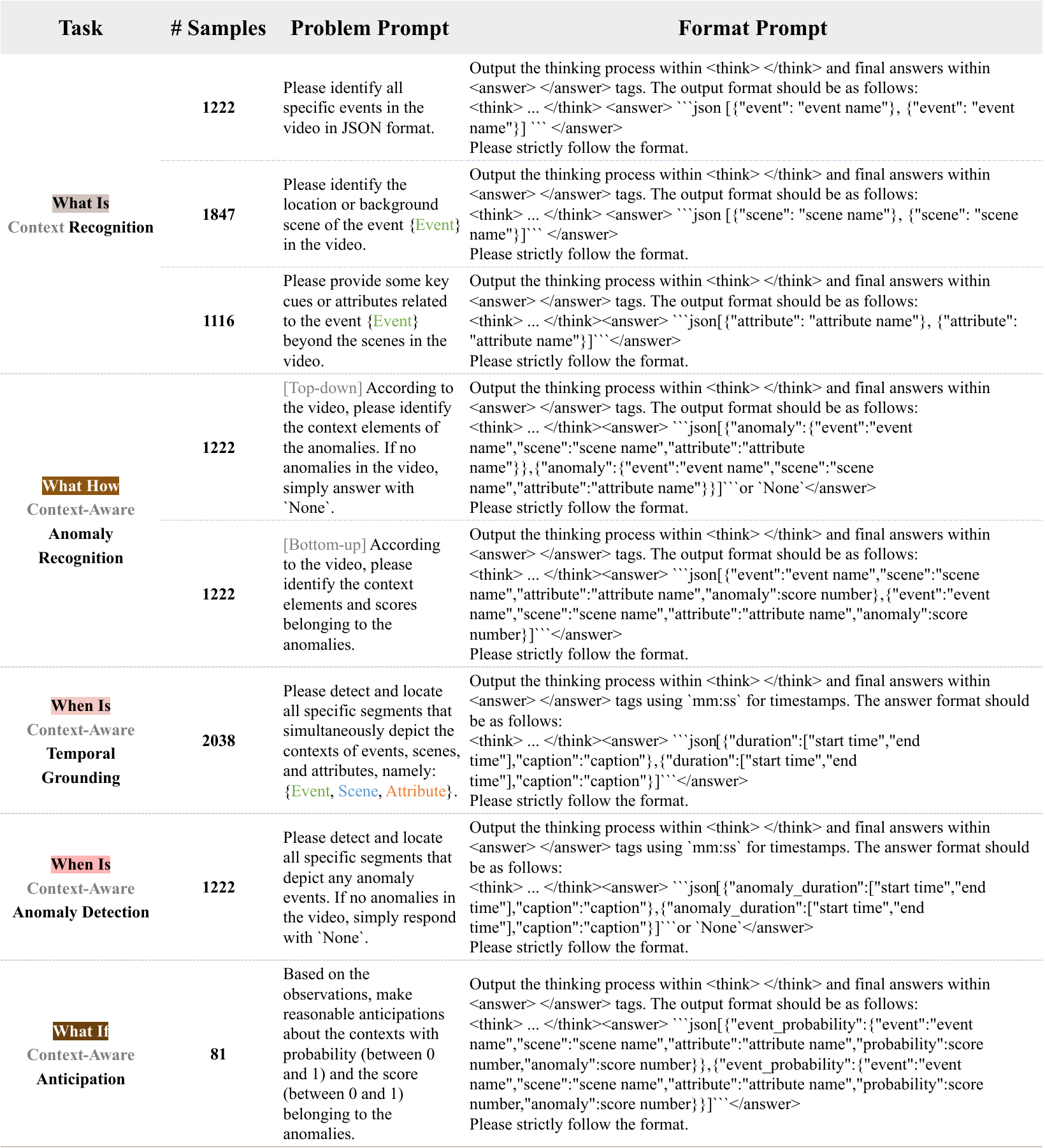}
    \caption{{Prompts for unified evaluation.} }
    \label{fig:prompt} 
\end{figure*}

The prompts used for the unified evaluation of \cuebench~are presented in~\figref{prompt}.
Each prompt starts with: ``\textit{This is a video showing some key events related to the safety, laws \& rules, or life \& health.}'' This is followed by the task-related problem prompt and the format prompt.
As mentioned in the main paper, the test set used in the unified evaluation consists of 1,222 videos, covering 1,249 unique anomalies and 194 unique normal context triplets.
Thus, the contexts and anomalies differ across videos, and the number of task samples varies accordingly.
Specifically, we formulate the context recognition task based on the context types present in each video and create context-aware anomaly recognition and detection tasks for all normal and abnormal cases.
And the context-aware temporal grounding task is generated based on the context triplets associated with each video.
For context-aware anticipation tasks, we select 81 surveillance videos captured by fixed surveillance cameras and trim them to retain only the segment containing the first event.
Surveillance footage often captures events in a linear and predictable manner, making it suitable for anticipation tasks.

\section{Additional Implementation Details}
\label{appen:implementation}
\subsection{Training Details}
We use LLaMA-Factory~\cite{zheng2024llamafactory} for efficient SFT, and then perform RFT following the GRPO parameter settings: number of completions $N=4$, temperature $\tau=0.9$, and the KL divergence ratio $\beta=0.04$.

\subsection{Separate Evaluation Details}
For event and anomaly recognition, we sample 8 frames from each segment containing the target events to assess the open-vocabulary generalization ability of specialized VLMs.
Anomaly recognition spans 1,443 categories, while event recognition includes 32 event types.
Anomaly detection is performed based on frame-level predictions: a frame is labeled as anomalous if its top-1 prediction falls within any anomaly class; otherwise, it is considered normal.
To evaluate the temporal grounding capabilities, we assess whether the model can accurately localize the start and end timestamps of target events within the video.

\section{Additional Experimental Results}
\label{appen:experiments}

\subsection{Additional Ablation Studies}
\label{appen:ablation}
\begin{table*}[!htbp]

\centering
\setlength{\tabcolsep}{1mm}
\resizebox{1\textwidth}{!}
{
\begin{tabular}{l|rrr|rr|rr|rrr|rrr|rr|rr|rrr}
\toprule
\multirow{2}{*}{Inference} & \multicolumn{3}{c|}{Event} & \multicolumn{2}{c|}{Scene} & \multicolumn{2}{c|}{Attribute} & \multicolumn{3}{c|}{Anomaly (TD)} & \multicolumn{3}{c|}{Anomaly (BU)} & \multicolumn{2}{c|}{Grounding} & \multicolumn{2}{c|}{Detection} & \multicolumn{3}{c}{Anticipation} \\
 & \multicolumn{1}{r}{Struct} & \multicolumn{1}{r}{Sem.} & \multicolumn{1}{r|}{Hier.} & \multicolumn{1}{c}{Struct} & \multicolumn{1}{c|}{Sem.} & \multicolumn{1}{c}{Struct} & \multicolumn{1}{c|}{Sem.} & \multicolumn{1}{c}{Struct} & \multicolumn{1}{c}{Sem.} & \multicolumn{1}{c|}{Hier.} & \multicolumn{1}{c}{Struct} & \multicolumn{1}{c}{Sem.} & \multicolumn{1}{c|}{Hier.} & \multicolumn{1}{c}{Struct} & \multicolumn{1}{c|}{TIoU} & \multicolumn{1}{c}{Struct} & \multicolumn{1}{c|}{TIoU} & \multicolumn{1}{c}{Struct} & \multicolumn{1}{c}{Sem.} & \multicolumn{1}{c}{Hier.} \\
 \midrule
 \rowcolor{gray!10} 
\multicolumn{21}{l}{\textbf{Baseline}} \\
 Direct & 59.3 & 34.9 & 15.2 & 68.1 & 43.4 & 57.3 & 37.1 & 54.4 & 37.8 & 1.4 & 62.2 & 31.8 & 2.0 & 43.5 & 16.8 & 63.1 & 19.2 & 65.5 & 3.3 & 0.4 \\
 CoT & 58.5 & 35.5 & 16.4 & 67.4 & 41.4 & 55.4 & 38.3 & 53.8 & 33.8 & 1.5 & 62.7 & 30.1 & 2.1 & 44.1 & 17.7 & 63.4 & 23.2 & 67.0 & 3.9 & 0.4 \\
\midrule
 \rowcolor{gray!10} 
\multicolumn{21}{l}{\textbf{SFT}} \\
 Direct & 82.3 & 72.8 & 46.1 & 95.4 & 81.6 & 78.9 & 66.0 & 65.3 & 60.1 & 6.5 & 80.0 & 58.2 & 7.1 & 55.9 & 35.0 & 52.0 & \textcolor{SkyBlue}{36.5} & 80.3 & 39.7 & 0.0 \\
CoT & {82.4} & {73.0} & {46.3} & 95.9 & 81.5 & 78.9 & {65.6} & 66.3 & 62.5 & {7.1} & {80.9} & {60.8} & {8.1} & 55.7 & {34.6} & 51.8 & \textcolor{Green}{\textbf{39.0}} & {80.6} & {39.1} & 0.0 \\
 \midrule
  \rowcolor{gray!10} 
\multicolumn{21}{l}{\textbf{RFT}} \\
Direct & 77.9 & 64.1 & 25.3 & 95.9 & 81.6 & 79.9 & 64.5 & 69.8 & 67.3 & 2.9 & 80.0 & 52.1 & 2.8 & 83.0 & 26.9 & 82.2 & 33.3 & 78.4 & 34.1 & {0.4} \\
CoT & 79.6 & 64.7 & 27.2 & {96.6} & \textcolor{Green}{\textbf{82.3}} & {80.8} & {65.0} & \textcolor{Green}{\textbf{72.0}} & {67.0} & 3.1 & 80.3 & 53.8 & 2.8 & {83.5} & 27.5 & \textcolor{Green}{\textbf{83.0}} & 34.9 & 80.0 & 35.5 & \textcolor{Green}{\textbf{0.6}} \\
 \midrule
\rowcolor{orange!10} 
\multicolumn{21}{l}{\textbf{\cue-R1}: $\lambda=0.5$ for RFT} \\
Direct & \textcolor{SkyBlue}{83.4} & 72.2 & 48.2 & 95.6 & 82.1 & 80.8 & \textcolor{SkyBlue}{67.8} & 71.2 & \textcolor{SkyBlue}{67.4} & 6.6 & 80.9 & 60.9 & 12.3 & 83.5 & 35.0 & 82.0 & 34.9 & 79.6 & 41.3 & 0.4 \\
CoT & \textcolor{Green}{\textbf{83.8}} & \textcolor{Green}{\textbf{73.6}} & 47.8 & 96.5 & 82.0 & 80.9 & 67.2 & \textcolor{Green}{\textbf{72.0}} & \textcolor{Green}{\textbf{68.1}} & 6.8 & 81.5 & 60.9 & 12.1 & \textcolor{Green}{83.8} & 35.3 & 82.2 & {35.3} & 80.6 & \textcolor{Green}{\textbf{43.9}} & 0.5 \\

\midrule
\rowcolor{red!10} 
\multicolumn{21}{l}{\textbf{\cue-R1}: $\lambda=0.2$ for RFT} \\
Direct & \textcolor{SkyBlue}{83.4} & \textcolor{SkyBlue}{72.9} & \textcolor{SkyBlue}{48.9} & \textcolor{SkyBlue}{96.5} & \textcolor{SkyBlue}{82.2} & \textcolor{SkyBlue}{81.0} & 67.7 & \textcolor{SkyBlue}{71.3} & \textcolor{SkyBlue}{67.4} & \textcolor{SkyBlue}{7.3} & \textcolor{SkyBlue}{\textbf{81.7}} & \textcolor{SkyBlue}{61.0} & \textcolor{SkyBlue}{12.8} & \textcolor{SkyBlue}{\textbf{83.9}} & \textcolor{SkyBlue}{\textbf{35.9}} & \textcolor{SkyBlue}{82.6} & 35.0 & \textcolor{SkyBlue}{80.4} & \textcolor{SkyBlue}{43.0} & \textcolor{SkyBlue}{0.5} \\
CoT & 83.7 & 73.2 & \textcolor{Green}{\textbf{49.2}} & \textcolor{Green}{\textbf{96.7}} & \textcolor{Green}{\textbf{82.3}} & \textcolor{Green}{\textbf{81.3}} & \textcolor{Green}{\textbf{68.1}} & 71.6 & 67.7 & \textcolor{Green}{\textbf{7.7}} & \textcolor{Green}{\textbf{81.7}} & \textcolor{Green}{\textbf{61.3}} & \textcolor{Green}{\textbf{13.6}} & \textcolor{Green}{83.8} & \textcolor{Green}{\textbf{35.9}} & 82.4 & 35.2 & \textcolor{Green}{\textbf{80.7}} & 43.7 & \textcolor{Green}{\textbf{0.6}} \\

\bottomrule
\end{tabular}
}
\caption{Ablation of different inference strategies and reward weighting ($\lambda$) across different training configurations based on Qwen2.5-VL-3B (Baseline). Inference is performed using the direct or CoT template.
The best \textcolor{Green}{\textbf{CoT}} and the best \textcolor{SkyBlue}{\textbf{direct}} inference results are colored, respectively. The best overall results are highlighted in \textbf{bold}.}
\label{tab:cot}
\end{table*}

We conduct an additional ablation study to examine the impacts of inference strategies and reward weighting ($\lambda$) across different training configurations, as shown in~\tabref{cot}. 
The results demonstrate that Chain-of-Thought (CoT) inference consistently outperforms direct inference, particularly in complex context-aware tasks \eg anomaly recognition, by enabling structured reasoning.
From the results, CoT improves TIoU and hierarchy scores across various settings, with the most notable improvement observed in SFT, where TIoU score of anomaly detection increases by 2.5\% ($36.5\rightarrow39.0$). 
For RFT with CoT inference, we find that a lower $\lambda$ value ($\lambda=0.2$), which places more emphasis on hierarchy rewards, achieves better alignment with the event and anomaly hierarchies without sacrificing semantic accuracy \ie improving the event hierarchy score by 1.4\% ($47.8\rightarrow49.2$) and the anomaly hierarchy score by 1.5\% ($12.1\rightarrow13.6$) over $\lambda=0.5$. 
The combined approach in \cue-R1 (with CoT, $\lambda=0.2$) delivers the best overall performance, highlighting the necessity of CoT reasoning, a hierarchy-aware reward design, and a hybrid SFT and RFT training strategy for effective and interpretable video anomaly understanding.

\subsection{Additional Qualitative Results}
\label{appen:qualitative}
\begin{figure*}[!t]
    \centering
    \includegraphics[width=\textwidth]{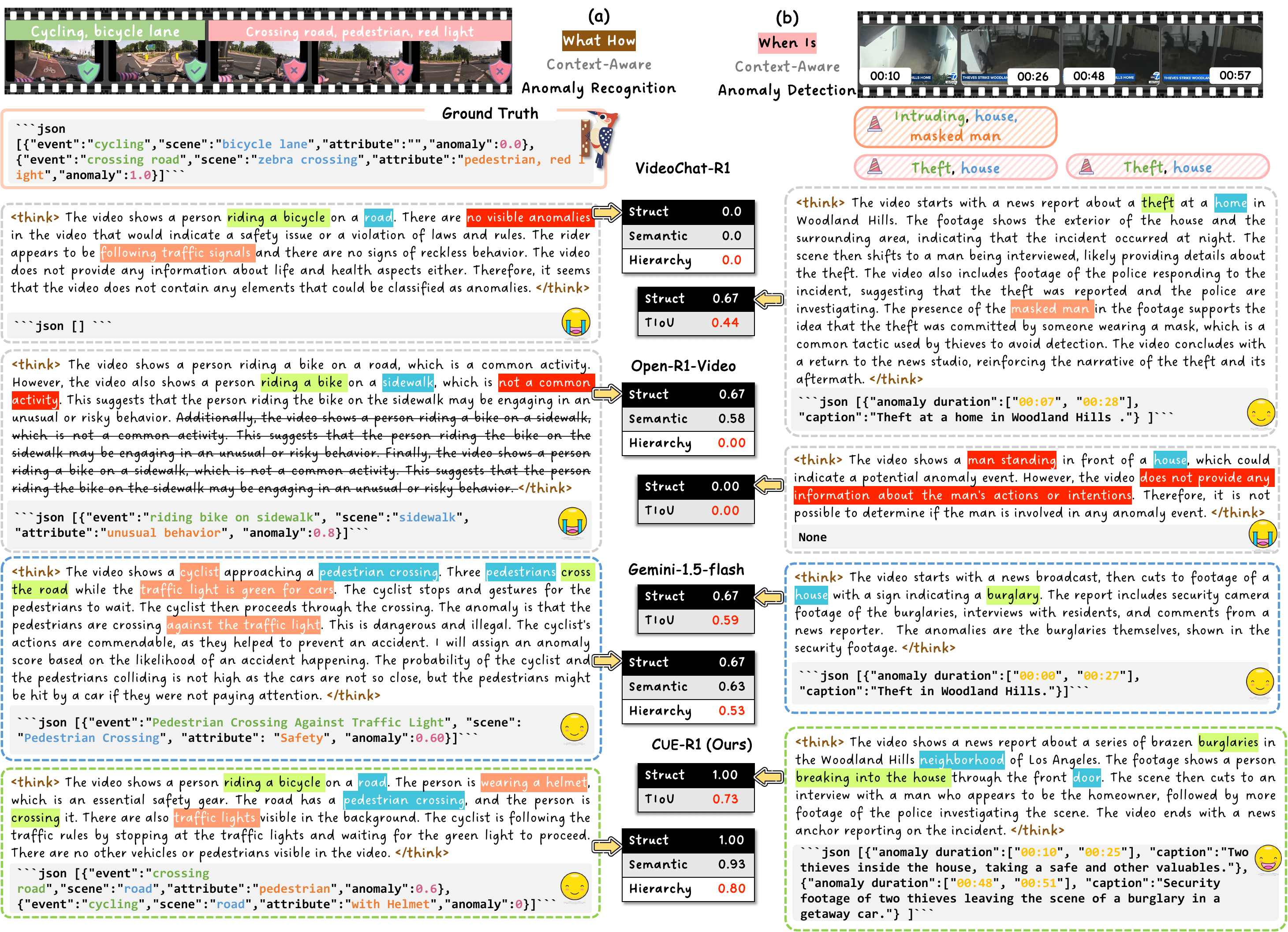}
    \caption{Qualitative comparison of context-aware anomaly recognition and detection with our \cue-R1 and other VLMs \ie VideoChat-R1, Open-R1-Video and Gemini-1.5-flash.}
    \label{fig:visual} 
\end{figure*}
\figref{visual} presents additional qualitative comparison of the generated reasoning process and final answers from various VLMs \ie VideoChat-R1~\cite{li2025videochat}, Open-R1-Video~\cite{wang-2025-open-r1-video}, Gemini-1.5-flash~\cite{team2024gemini}, and our \cue-R1. 
Our \cue-R1, incorporating structured output and hierarchical alignment, allows precise disambiguation between conditional normalities and anomalies, enabling trustworthy context-aware VAU.

\section{Discussions}
\label{appen:discussions}

\subsection{Related Works}
\label{appen:related}
\paragraph{Vision-Language Models.}
The advent of large vision-language models (VLMs) has spurred research in multi-modal contexts, especially in video understanding. With seminal works such as CLIP~\cite{yu2025learning} serving as \textit{discriminative VLMs}, it has shed light on efficient vision-language alignment for open-vocabulary video applications in action recognition~\cite{yu2025learning,weng2023open,huang2024froster}, action detection~\cite{wang2022beyond}, action anticipation~\cite{zhao2023antgpt}, anomaly detection~\cite{wu2024open,wu2024vadclip} and etc. 
Recent research interest has shifted towards the development of \textit{generative VLMs} with powerful large language models (LLMs)~\cite{jaech2024openai} to thrive on multi-modal understanding and generation capabilities on captioning~\cite{li2022blip,li2023blip} and question-answering~\cite{zhu2025internvl3}, as exemplified by a series of groundbreaking commercial models like GPT-4o~\cite{hurst2024gpt}, Gemini-1.5~\cite{team2024gemini}.
The emergence of open-source LLMs~\cite{thawakar2025llamav}, coupled with progress in multi-modal alignment for vision encoding, has enabled significant research into public generative VLMs, breeding impressive works such as VideoChat~\cite{li2025videochat}, Video-LLaMA~\cite{zhang2025videollama}, Qwen-VL~\cite{bai2025qwen2} and InternVideo~\cite{wang2025internvideo2}.
With significant achievements in general video understanding, such progress underscores the need to gauge the capabilities of these VLMs for unified VAU in real-world scenarios.

\paragraph{Video Anomaly Understanding.}
Traditional VAU works primarily focus on VAD settings, with most benchmarks typically restricted to simulate anomalies in the wild~\cite{acsintoae2022ubnormal} or to study anomalies within limited real-world scenarios~\cite{aung2025multi,ramachandra2020street,adam2008robust}.
In this direction, VAD approaches have emerged to detect deviations from the learned normal patterns in a semi~\cite{cao2023new} or weakly-supervised~\cite{sultani2018real} manner, since anomalous events are scattered and rare occurrences in practice.
Despite the exploration of semantic information in open-vocabulary VAD~\cite{wu2024open} or scene-dependencies~\cite{cao2025scene} underlying anomalies in NWPU Campus~\cite{cao2023new}, a significant gap remains in real-world anomaly understanding with context indispensability~\cite{zhu2024advancing}.
Building upon the significant achievements of VLMs, the research community has pursued further exploration of VAU benchmarks to probe the multi-modal anomaly understanding abilities with VQA setups of captioning~\cite{zhang2025holmes}, anomaly reasoning~\cite{du2024uncovering,huang2025vad} and user interactions~\cite{tang2024hawk}.
Although these benchmarks have enriched anomaly dependency from superficial deviations to anomalous events in open-world multi-modalities, they still lag behind the broader understanding of diverse context-dependent anomalies and normalities in real-world.
In response to this need, our {\cuebench}~is the first dedicated to context-aware video anomalies in real-world within a unified evaluation across diverse tasks, marking a significant stride toward more nuanced and comprehensive VAU. 
It also features a comprehensive hierarchical taxonomy for absolute and conditional anomalies or normalities, tapping into the underlying refined semantics of scenes and attributes.

\paragraph{Video Anomaly Understanding with Generative Vision-Language Models.}
%
Recent studies have demonstrated the reliable capabilities of generative vision-language models in VAU. 
Leveraging the exceptional logical reasoning capabilities of VLMs, A-Guardian~\cite{du2024uncovering} introduces a prompt mechanism to guide VLMs to focus on critical anomaly clues in the video, thus building a logic chain of the cause-effect.
HAWK~\cite{tang2024hawk} pushes VLM's motion-related interpretation capability of video anomalies by incorporating the motion modality via motion attention reinforcement.
Recently, Holmes-VAU~\cite{zhang2025holmes} proposes the anomaly-focused temporal sampler to reduce the temporal redundancy of VLMs for accurate anomaly detection and understanding in long-term videos.
Unlike these works, we delve into the GRPO-based reinforcement learning from Open-R1~\cite{guo2025deepseek} and adopt verifiable rewards to perform post-training, enhancing the VLMs' reasoning capabilities to address the challenges within \cuebench.

\subsection{Broader Impacts}
\label{appen:impacts}
The introduction of \cuebench~and \cue-R1 carries significant implications for real-world VAU applications, particularly in safety-critical domains. 
By advancing context-awareness, this work enables more nuanced AI systems for surveillance, public safety, and autonomous monitoring. 
For instance, in smart cities, robust VAU could improve traffic incident response or hazard detection in crowded spaces. 
In healthcare, it could assist in fall detection for personal care, where context dictates anomaly severity.
However, these advancements also raise ethical concerns. 
The benchmark’s reliance on YouTube-sourced data may inherit societal biases (\eg cultural norms for ``normal'' behavior), potentially leading to over-policing of marginalized groups. 

\subsection{Limitations and Future Work}
\label{appen:limitations}
Despite its contributions, our work faces several limitations:

\paragraph{Dataset Scope.} 
While CueBench includes 174 scenes and 198 attributes, its coverage of real-world anomaly diversity is inherently limited. The current 14 conditional events omit critical environmental contexts (\eg lighting or weather), and reliance on public YouTube sources introduces potential geographic and cultural biases.

\paragraph{Annotation Challenges.}
Data construction is constrained by manual efforts, as existing tools fail to capture nuanced multi-modal contexts. 
The dataset also lacks free-text rationales and captions, hindering CoT's distillation for explainable reasoning.

\paragraph{Evaluation Gaps.}
Our unified framework prioritizes structured outputs but may undervalue free-form reasoning. The hierarchical scoring mechanism, though semantically grounded, can misalign with human judgment in ambiguous cases.

\paragraph{Generalization.}
\cue-R1 excels on \cuebench~in the open-world settings.
However, its event-centric anomaly understanding limits adaptability to anomalies defined solely by unexpected objects or visual cues (e.g., an unattended bag in a secure area), which requires non-event-centric reasoning.

Future work will address these by expanding dataset diversity with an advanced automatic pipeline, developing lightweight models, integrating human-in-the-loop evaluation for ambiguous cases, and exploring hybrid evaluation to balance structure with open-ended reasoning.


\end{document}